\documentclass{article}
\usepackage{ijcai23}

\usepackage{times}
\usepackage{soul}
\usepackage{url}
\usepackage[hidelinks]{hyperref}
\usepackage[utf8]{inputenc}
\usepackage[small]{caption}
\usepackage{graphicx}
\usepackage{amsmath}
\usepackage{amsthm}
\usepackage{booktabs}
\usepackage{algorithm}
\usepackage{algorithmic}
\usepackage[switch]{lineno}
\usepackage{multirow}

\usepackage{color}
\usepackage{amssymb}
\usepackage{pifont}
\newcommand{\cmark}{\ding{51}}%
\newcommand{\xmark}{\ding{55}}%

\usepackage{mathtools}
\usepackage{xspace}

\title{Efficiency 360: Efficient Vision Transformers}

\author{
Badri N. Patro$^1$
\and
Vijay Srinivas Agneeswaran$^2$
\affiliations
$^1$$^2$Microsoft\\
\emails
\{badripatro, vagneeswaran\}@microsoft.com
}

\begin{document}

\maketitle

\begin{abstract}

    Transformers are widely used for solving tasks in natural language processing, computer vision, speech, and music domains. In this paper, we talk about the efficiency of transformers in terms of memory (the number of parameters), computation cost (number of floating points operations),  and performance of models, including accuracy, robustness of the model, and fair \& bias-free features. We mainly discuss the vision transformer for the image classification task. Our contribution is to introduce an efficient 360 framework, which includes various aspects of the vision transformer, to make it more efficient for industrial applications. By considering those applications, we categorize them into multiple dimensions such as privacy, robustness, transparency, fairness, inclusiveness, continual learning, probabilistic models, approximation, computational complexity, and spectral complexity. We compare various vision transformer models based on their performance, the number of parameters, and the number of floating point operations (FLOPs) on multiple datasets.
\end{abstract}

\section{Introduction}
\label{sec:intro}
  \begin{figure}[h]%
\centering
\includegraphics[width=0.49\textwidth]{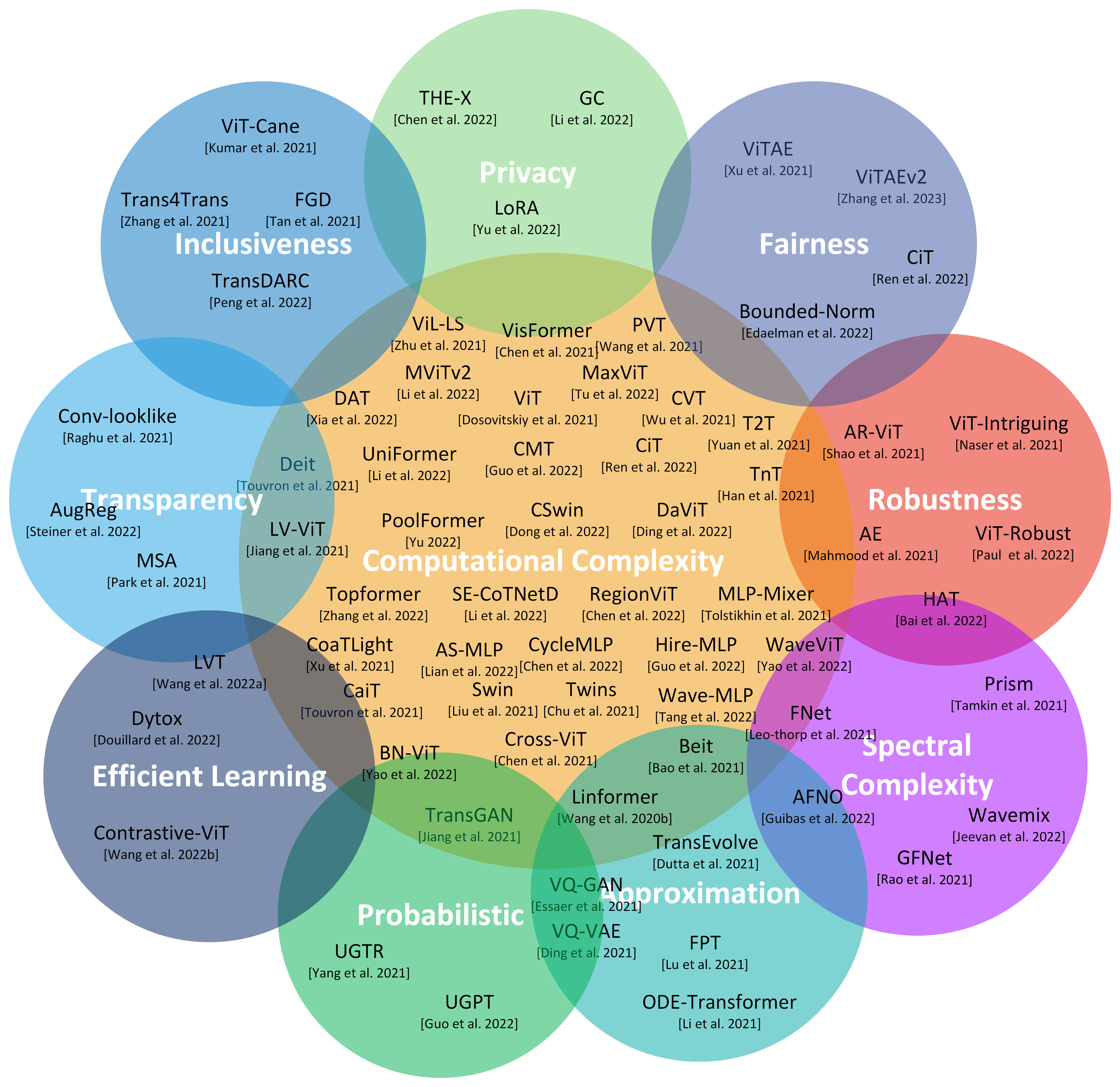}

\vspace{-0.4em}
\caption{Efficient Vision  Transformers: Venn diagram of the efficient transformer models (Efficiency-360). This includes the robustness of a model, privacy of the model,  bias and fairness, transparency, inclusiveness, efficient learning, probabilistic models, model approximations, the computational complexity of a model, and spectral complexity techniques}\label{fig1}
\vspace{-1.2em}
\end{figure}

Transformers such as attention all you need ~\cite{vaswani2017attention}and Bidirectional Encoder Representations from Transformers (BERT)~\cite{kenton2019bert} have recently become popular in the Machine Learning world for Natural Language Processing (NLP) tasks such as machine translation, text summarization, question answering, protein fold prediction, and even image processing tasks. ChatGPT~\cite{gpt} based transformer model and other large language models have garnered public attention in the last few weeks. They are used to assist humans in various ways, including answering questions, generating articles, and even as a coding assistant, which is helpful for both industrial applications and academia. This work focuses on Vision transformers and their application in various industrial dimensions. A few dimensions that are important from an industry perspective of these advanced models include robustness, privacy, transparency, inclusiveness, and continual and distributed learning. 

\begin{table*}[htb]
\centering
\begin{tabular}{p{15em}p{5em}p{5em}p{5em}p{5em}p{5em}p{4em}p{4em}p{2em}} 
\hline
Model & Position Embedding & Attention Type  & Network Architecture  & Attention Network & Structure & Extra \qquad Label \\
\hline\hline
ViT~\cite{dosoViTskiy2020image} & \textcolor{green}{\cmark} & Global  & MSA & Linear & Isotropic&  \textcolor{red}{\xmark}\\
DeiT~\cite{touvron2021training} & \textcolor{green}{\cmark} & Global  & MSA & Linear & Isotropic&  \textcolor{red}{\xmark}\\

TNT~\cite{han2021transformer} & \textcolor{green}{\cmark} & Global  & MSA & Linear & Isotropic&  \textcolor{red}{\xmark}\\
T2T~\cite{yuan2021tokens} & \textcolor{green}{\cmark} & Global  & MSA & Linear & Isotropic&  \textcolor{red}{\xmark}\\
Cross-ViT~\cite{chen2021crossViT} & \textcolor{green}{\cmark} & Global  & MSA & Linear & Isotropic&  \textcolor{red}{\xmark}\\
PVT~\cite{wang2021pyramid} & \textcolor{green}{\cmark} & Global& MSA & Linear & Pyramid &  \textcolor{red}{\xmark}\\
Swin~\cite{liu2021swin} & \textcolor{green}{\cmark} & Local  & MSA & Linear & Pyramid &  \textcolor{red}{\xmark}\\
Twin~\cite{chu2021twins} & \textcolor{green}{\cmark} & LG  & MSA & Linear & Pyramid &  \textcolor{red}{\xmark}\\
CiT~\cite{ren2022co} & \textcolor{green}{\cmark} & Global  & MSA & Linear & Isotropic&  \textcolor{red}{\xmark}\\
CSwin~\cite{dong2022cswin} & \textcolor{green}{\cmark} & LG  & MSA & Linear & Pyramid&  \textcolor{red}{\xmark}\\
LV-ViT~\cite{jiang2021all} & \textcolor{green}{\cmark} & Global  & MSA & Linear & Isotropic&  \textcolor{green}{\cmark}\\
WaveViT~\cite{yao2022wave} & \textcolor{green}{\cmark} & Global  & MSA & Linear & Isotropic&  \textcolor{green}{\cmark}\\
CMT~\cite{guo2022cmt} & \textcolor{green}{\cmark} & LG  & MSA & Linear & Isotropic&  \textcolor{red}{\xmark}\\

RegionViT~\cite{chen2022regionViT} & \textcolor{green}{\cmark} & LR  & MSA & Linear & Pyramid&  \textcolor{red}{\xmark}\\
CoaT~\cite{xu2021co} & \textcolor{green}{\cmark} & Local  & MSA & Linear & Pyramid&  \textcolor{red}{\xmark}\\

CvT~\cite{wu2021CvT} & \textcolor{red}{\xmark} &  LG   & MSA & Convolution & Pyramid&  \textcolor{red}{\xmark}\\
UniFormer~\cite{li2022uniformer} & \textcolor{green}{\cmark} & LG  & MHRA & Linear & Pyramid&  \textcolor{red}{\xmark}\\

MLP-Mixer~\cite{tolstikhin2021mlp} & \textcolor{green}{\cmark} & Global*  & Mixer &   \textcolor{red}{\xmark} & Isotropic&  \textcolor{red}{\xmark}\\

AS-MLP  ~\cite{lianmlp} & \textcolor{green}{\cmark} & Local*  & Mixer &   \textcolor{red}{\xmark} & Isotropic&  \textcolor{red}{\xmark}\\

CycleMLP~\cite{chencyclemlp}  & \textcolor{green}{\cmark} & Global*  & Mixer &   \textcolor{red}{\xmark} & Isotropic&  \textcolor{red}{\xmark}\\

PoolFormer ~\cite{yu2022metaformer} & \textcolor{green}{\cmark} & Global*  & Pooler &   \textcolor{red}{\xmark} & Isotropic&  \textcolor{red}{\xmark}\\

GFNet~\cite{rao2021global} & \textcolor{green}{\cmark} & Global*   & SGN &  \textcolor{red}{\xmark} & Isotropic&  \textcolor{red}{\xmark}\\

GFNet-H~\cite{rao2021global} & \textcolor{green}{\cmark} & Global*  & SGN & \textcolor{red}{\xmark} & Pyramid&  \textcolor{red}{\xmark}\\

AFNO~\cite{guibas2021efficient} & \textcolor{green}{\cmark} & Global*  & SGN &  \textcolor{red}{\xmark} & Isotropic&  \textcolor{red}{\xmark}\\

\hline

\hline
\end{tabular}
\captionof{table}{ We compare transformer models based on position Embedding, token embedding, network Architecture, attention network, Hierarchical Type, and extra labels.  to comparison. SGN stands to be a Spectral Gating Network, and MSA stands for a Multi-headed Self Attention network. MHRA stands for Multi-Head Relation Aggregator. LG stands for Local + Global. LR stands for Local + Regional. "*" indicate that it's not an attention-type network. }\label{EVT_tab_0}
\end{table*}

In literature, Tay et al.~\cite{tay2020efficient} have discussed efficient transformers primarily by considering natural language processing-based transformers. In this article, we build on a survey of efficient transformers in the vision domain, which has a different characterization compared to the existing surveys. We include more recent work on advanced transformers (especially those published in 2021 and 2022) in our current survey. Interesting research directions open up as a result, which we discuss in later sections of this article. As we shall be discussing, the survey opens up research in inclusiveness and privacy. It also suggests that the advanced transformer models work on high-resolution data, opening up research in climate modeling and oceanography (wave-breaking kind of applications). The other contribution of this paper is the efficient 360 framework, which helps provide a holistic view of transformers across these dimensions. 
\begin{figure}[ht]%
\centering
\includegraphics[width=0.49\textwidth]{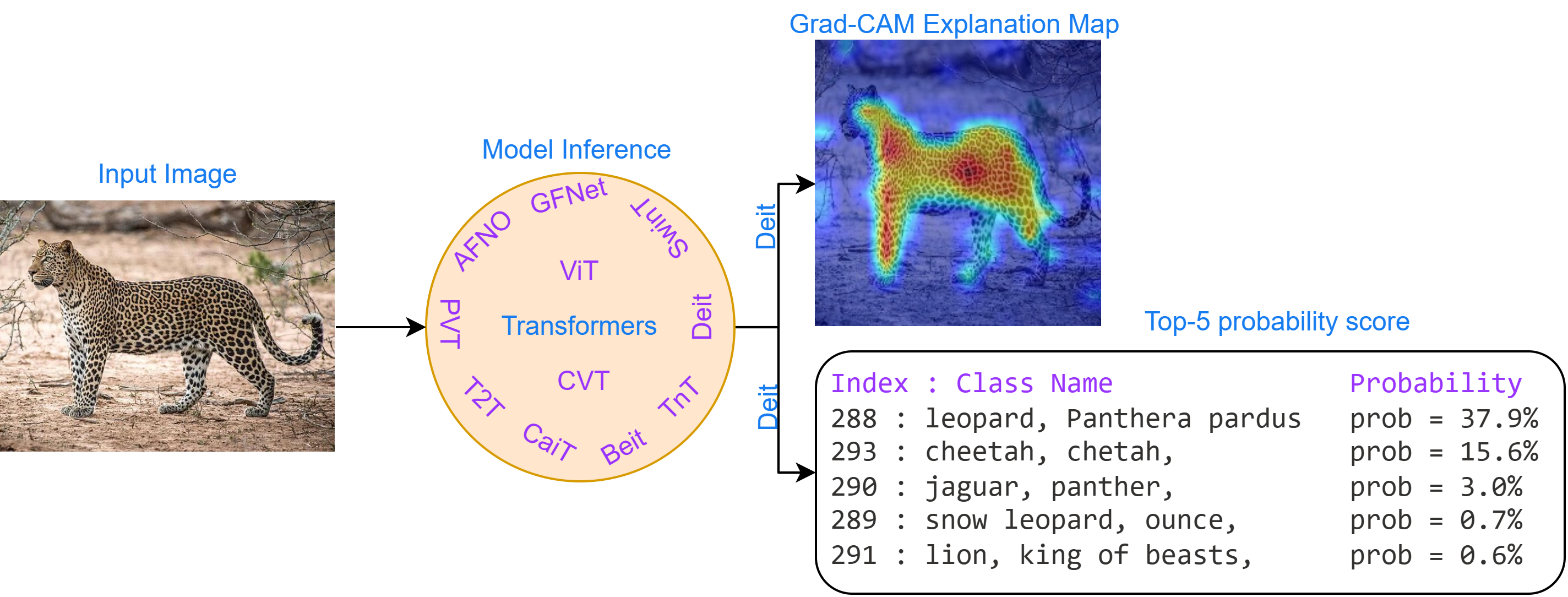}
\caption{Inference on efficient transformer model (DeiT). }\label{fig2}
\end{figure}

\begin{figure*}
\centering
\includegraphics[width=0.99\textwidth]{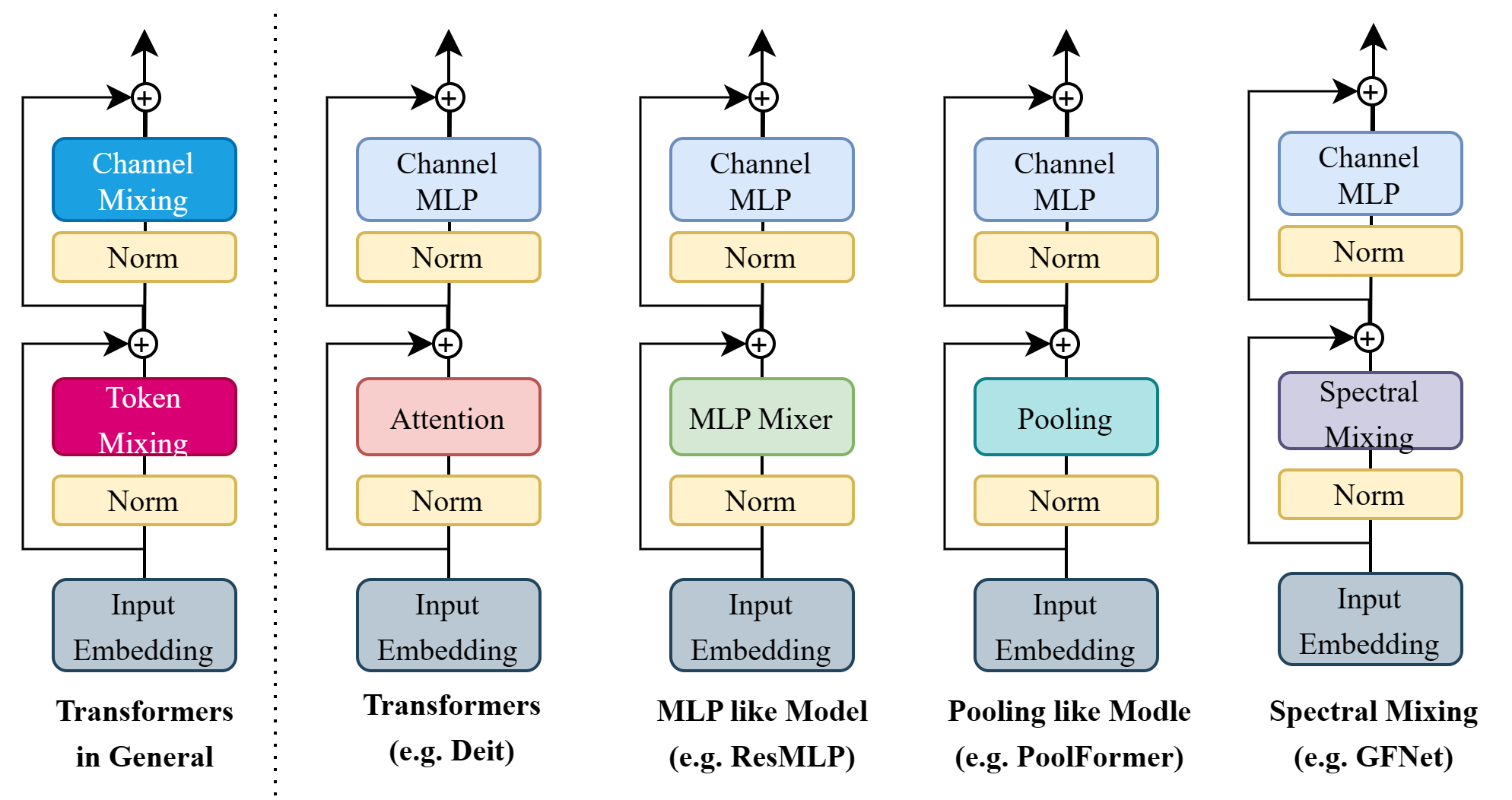}
\caption{This figure shows a comparison of various transformer model architectures. In general, the transformer architecture divides two parts one token mixing and another one channel mixing. We show various token-mixing model architectures. It is an extended version of the diagram with spectral mixing shown in  PoolFormer \protect\cite{yu2022metaformer}.}\label{fig_trans}
\vspace{-0.7em}
\end{figure*}
In this work, we start our survey by considering a classification task. The transformer model needs to classify a given image into  1000 pre-defined classes. To benchmark the model performance, the research community chooses the ImageNet-1K~\cite{deng2009imagenet} dataset. We provide sample examples of the given image of a leopard and classify it into its classes as shown in figure-\ref{fig2}. We select DeiT-base~\cite{touvron2021training} as a pre-train transformer model and obtain its prediction index \& probability scores. We visualize the Grad-CAM-based explanation for the prediction. We start with a discussion of how efficient DeiT~\cite{touvron2021training} is over ViT~\cite{dosoViTskiy2020image}. Here we define efficiency in terms of model parameters, computational cost ( i.e., training, inference time,) and performance (Top-1 accuracy). Along with this, we discuss various dimensions for efficient models like bias-free models,  robust features, privacy in the model, transparency in model, efficient way of learning, easily deploy-able (inclusiveness), and efficient architecture(self-Attention, MLP-Mixer, and Spectral models).

 In the figure-\ref{fig1}, we have included the major categories of efficient transformers: computational complexity, spectral complexity, robustness, privacy, approximation, efficient learning, transparency, fairness, and inclusiveness.  We review each in turn in the subsequent sections.  Due to the page constraint, our high-resolution figures, plots, and details comparison of state-of-the-art models are available on our GitHub page. \footnote{https://github.com/badripatro/efficient360} 


We compare transformer models based on the position Embedding, token embedding, network Architecture, attention network, Hierarchical Type, and extra labels in table-\ref{EVT_tab_0}. We check whether position embedding is used in the transformer or not and the type of token embedding in the transformer model, like overlapping or non-overlapping. We consider various types of the core architecture of the transformer models like Multi-Headed Self Attention, MLP-mixer-based architecture, and Spectral Gating Networks( using learnable filter weights). Also, we check with types of attention networks like linear layer or Convolution Neural Networks for QKV. Hierarchical architecture is used in the transformer model. We also compare with models that used extra data for training like Wave-ViT~\cite{yao2022wave} and LV-ViT~\cite{jiang2021all}. SGN tends to Spectral Gating Network, and MSA tends to a Multi-headed Self Attention network.

Figure-\ref{fig_trans} shows the general architecture of transformer models in the vision domain. It contains two main parts Token mixing and channel mixing. The figure shows various efficient token-mixing techniques, Such as attention-type token mixing, MLP-mixture-based token mixing, pooling-based token mixing, and spectral mixing techniques. The channel mixing techniques are also similar for all types of transformers.

\section{Efficient 360 Framework} 
The growing size of the neural networks results in improved model performance. As model size increases, which results in an increase in memory consumption and computational requirements (for storing weights, activations, and gradients) while training, those models also increase. Now the challenge is, how do we design efficient transformer models for a visual domain?

\subsection{Computational complexity} 

These transformers address the $O(N^2)$ computational complexity of transformers in various ways. One of the critical issues in a transformer is its quadratic complexity concerning the input sequence length—along dimensions relating to both computation and memory. The implication is that one has to compute the $N\times N$ attention matrix for every layer and attention head. Various approaches have been tried to reduce this $O(N^2)$ complexity, including the use of caching architectures. The sparse transformer is one of the popular methods to address this complexity. Each output position computes weights from a subset of input positions. If the subset is $\sqrt(N)$, then the complexity of the transformer reduces to $O(N \times  \sqrt(N))$, allowing it to handle long-range dependencies.  

We start with Vision Transformer (ViT)~\cite{dosoViTskiy2020image} model considers the image a 16 x16 word and is used to classify the image into predefined categories. In the ViT model, each image is split into a sequence of tokens of fixed length and then applied to multiple transformer layers to capture the global relationship across the token for the classification task. However, the performance of ViT is lower than CNNs on the ImageNet dataset when we train from scratch. Yuan et al.~\cite{yuan2021tokens} find the reason for the performance degradation of the ViT model as 1. The importance of local structures, such as lines and edges among the neighboring pixels, is not captured by the simple tokenization of the input image. This leads to low training sample efficiency. 2. The redundancy of the attention module of the ViT leads to limited feature richness and limited training samples for a fixed computation budget. To overcome these issues of the ViT model, Yuan et al.~\cite{yuan2021tokens} proposed a Tokens-to-token ViT method for training vision transformers from scratch on ImageNet. The proposed method includes layer-wise tokens to the token transformation that structures the image to the token by recursively aggregating neighboring tokens to one token (Tokens-to-token). This captures the local structure represented by the surrounding tokens, and the token's length is also reduced. The method also includes a deep narrow structure as an efficient backbone for the vision transformer. Transformer iN Transformer (TNT)~\cite{han2021transformer} finds an issue in the local patch to capture color information and complex object details in the vision transformer. The author has proposed the TNT method, which considers 16x16 words as a visual sentence and further divided into small patches of 4x4 as the visual words. This makes boosts transformer performance by 1.7\% compared to the state-of-the-art method.

Touvron et al. ~\cite{touvron2021training} proposed an efficient transformer model based on distillation technique (DeiT). It uses a teacher-student strategy that relies on a distillation token to ensure that a student learns from a teacher through attention. Bao et al.~\cite{bao2021beit} have proposed a masked image model task for a pretrained vision transformer. The author proposes a self-supervision–based vision representation model, Bidirectional Encoder representation from Image Transformers (BEiT), which follows the BERT ~\cite{kenton2019bert} method developed for the Natural Language Processing area. In this method, each image is considered as two views: One of the image patches of size 16 x 16 pixels and the other of discrete visual tokens. The original image is tokenized into visual tokens, with some image patches randomly masked and then fed to the backbone pre-trained transformer. After training BEiT, the model can be fine-tuned for the downstream tasks. 

\begin{figure}
\centering
\includegraphics[width=0.499\textwidth]{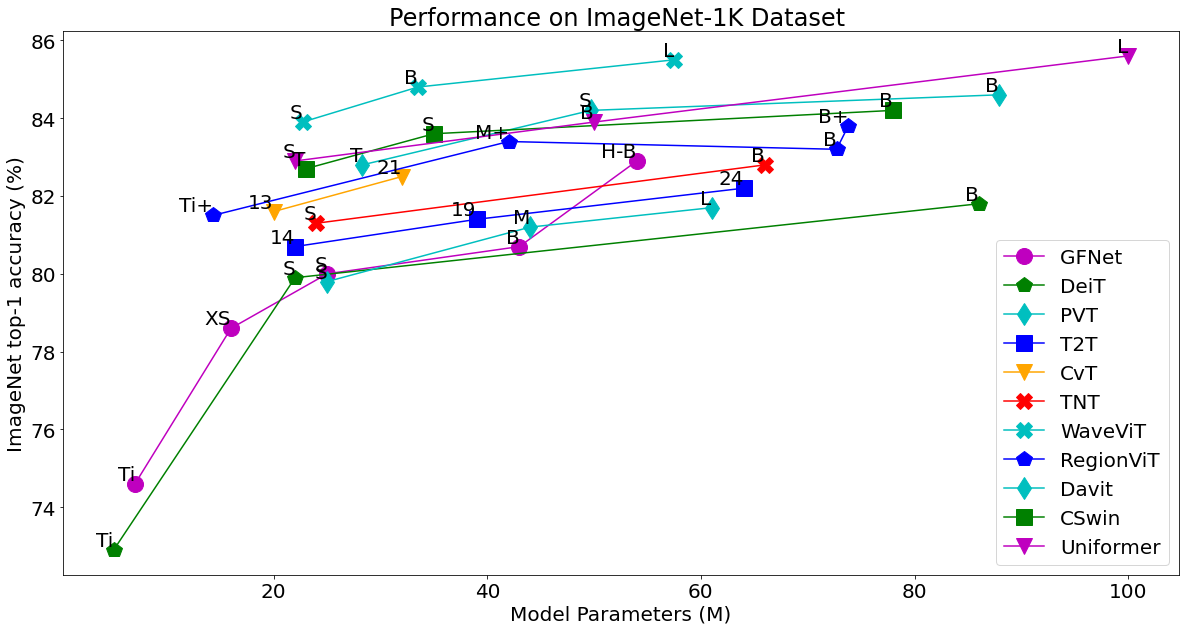}
\caption{This figure shows the performance of Various models across different model architectures like Tiny (T), Small (S), Base (B), and Large (L). This plot shows the variation of top-1 accuracy based on the various architectures like T, S, B, and L. We plot a number of parameters(M) for each architecture 
 vs. its top-1 performance.}\label{fig3}
\vspace{-0.7em}
\end{figure}

\begin{figure*}
\centering
\includegraphics[width=0.99\textwidth]{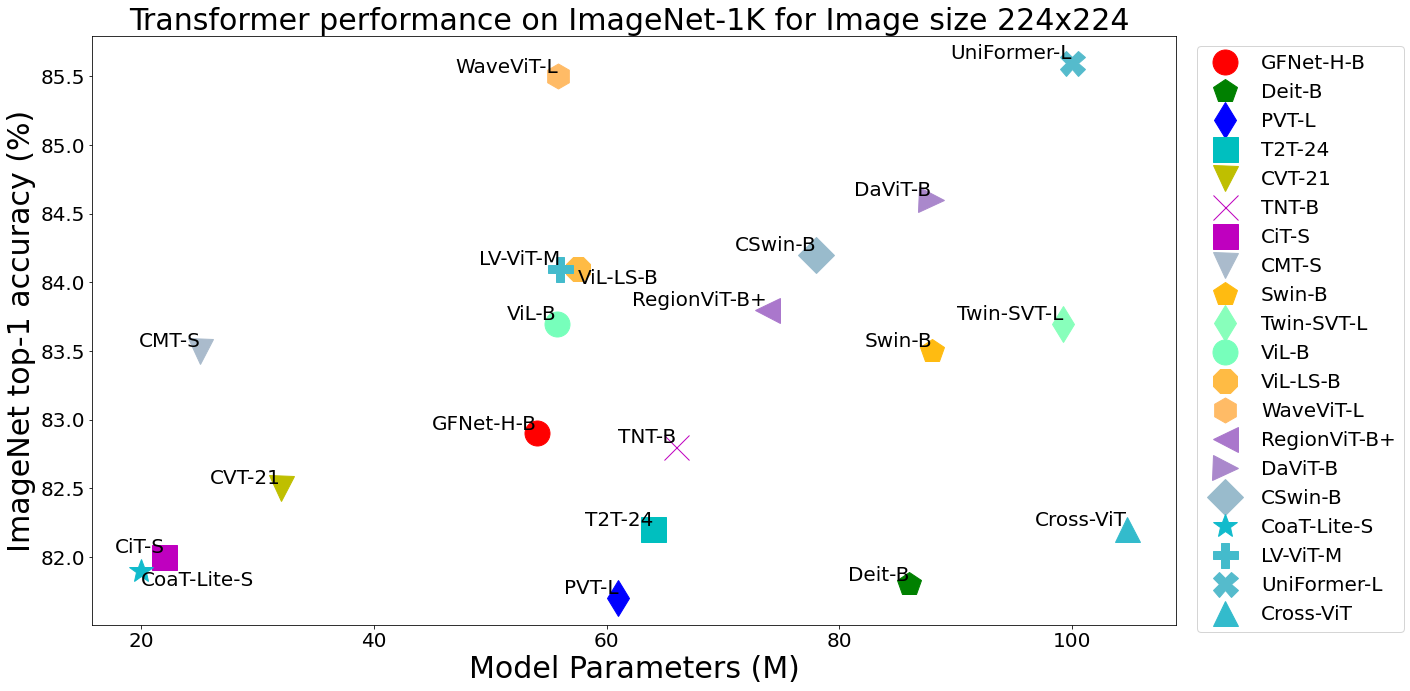}
\caption{This figure shows the performance of various state-of-the-art vision transformer models across a number of parameters. We plot the number of parameters(M) of different transformer models vs. their top-1 performance on the ImageNet-1K dataset. From this figure, we conclude that those models which are towards \textbf{top-left} are the best efficient models. For example WaveViT-L\~cite{yao2022wave} is the best efficient model.}\label{fig4}
\vspace{-0.7em}
\end{figure*}
Touvron et al.~\cite{touvron2021going} have studied architectural optimization in the image transformer model. The author builds a Class attention image Transformer(CaiT) and optimizes a deeper transformer network for the image classification task. The CaiT method has two main contributions; one is LayerScale, which is the multiplication of a diagonal matrix with one output of each residual block, and the second one is Class attention, which is a set of layers that compile a collection of patch embedding into class embedding and fed to the linear classifier.

Chen et al. have proposed a transformer model Cross-ViT~\cite{chen2021crossViT} to combine image patches, i.e., tokens in transformer,  of different sizes to produce strong feature representation using dual branch transformers. One transformer branch handles small patch tokens, and the other transformer handles large patch tokens with two separate branches of different computational complexity. Single token for each branch as a query to exchange information with other branches using effective token fusing module on cross attention. It is efficient in terms of a number of FLOPS and model parameters.

Wang et al. ~\cite{wang2021pyramid} have proposed a Pyramid vision Transformer (PVT) for dense prediction without convolutions. Vision-based transformers encounter difficulties while porting these transformers to dense prediction tasks. The PVT overcomes this issue. PVT is helpful for pixel-level dense predictions without convolution and non-maximal suppression, such as object detection methods. It is easy to port transformers using progressive shrinking pyramids and spatial reduction attention. Finally, PVT is evaluated on various tasks such as image classification, object detection, instance, and semantic segmentation tasks. Liu et al. ~\cite{liu2021swin} have discussed the issue of adapting transformers from the language domain to the visual domain in ways that encompass a large variance of visual entities and high-resolution pixels of images compared to words in the text. To address this issue, the author proposed Swin Transformer ~\cite{liu2021swin}, a hierarchical transformer method whose representation is computed using shifted windows. This technique overcomes the issue of non-overlapping local windows of self-attention more efficiently.

Transformer models provide high representation power compare to CNN-based models. But these models are not good at dense prediction due to excessive memory, computational cost, and feature, which can influence the irreverent part of the image. To avoid these issues, the sparse attention method proposed in PVT~\cite{wang2021pyramid} and Swin~\cite{liu2021swin} transformers are data agnostic and limit the long-range relationship. To solve this issue deformable self-attention method is proposed, where the position of the value and key pair is selected in a data-dependent way. The technique is called Deformable Attention Transformer (DAT)~\cite{xia2022vision}.

Chu et al. ~\cite{chu2021twins} have discussed the importance of spatial attention for success in the transformer's performance on various tasks. The authors proposed two simple and efficient architectures such as Twins-PCPVT and Twins-SVT. This paper uses a separable depth-wise convolution attention mechanism known as spatial-separable self-attention (SSSA). SSSA uses two types of attention operations: Locally grouped self-attention (LSA) and globally sub-sampled attention (GSA). LSA deals with fine-grained and short-distance information, while GSA deals with long-distance sequences and global information. The second proposed method, Twins-SVT, uses LSA and GSA with matrix multiplication. The author compares Twins-PCPVT with the similar architecture PVT~\cite{wang2021pyramid}, and Twins-SVT with similar architecture Swin ~\cite{liu2021swin} transformer. This makes it more efficient in terms of performance.

 CvT~\cite{wu2021CvT} is an efficient transformer model which introduces convolution token embedding and convolution transformer block in ViT architecture.
 The Convolution Neural Networks are good at shift, scale, and distortion invariant to ViT architecture, which maintains the merits of transformer architecture by using dynamic attention, global context, and better generalization. CvT gets all the benefits from CNNs, like a local receptive field, shared weights, and spatial sub-sampling, along with keeping all the advantages of transformer models. It improves performance compared to CNNs-based models ResNet~\cite{he2016deep} and transformer-based models ViT~\cite{dosoViTskiy2020image} and DeiT~\cite{touvron2021training}. CvT is also more efficient regarding the number of FLOPS and parameters. We have analyzed model architecture-wise comparison for CvT, DeiT, PVT, TNT, and T2T transformer models as shown in the figure-\ref{fig3}.
 
 Xu et al. have proposed a transformer model CoaT ~\cite{xu2021co} for image classification tasks trained with co-scale and conv-attention mechanisms. The co-scale mechanism allows learning representation at different scales to communicate effectively with each other. The conv-attention mechanism realizes relative position embedding with convolutions in the factorized attention module, which is computationally efficient. Gua et al. have proposed a CMT~\cite{guo2022cmt} transformer based on a hybrid network, which has the advantage of the transformer capturing long-range dependency and convolution neural network to extract local information. It is efficient in the number of FLOPS and the network's performance.

\begin{table*}[]
\centering
\begin{tabular}{p{3em}|p{17em}p{3em}p{3em}p{3em}p{4em}p{4em}p{2em}}
\hline
Image Size & Method & Network Type  & \#Param (M) & FLOPS (G) & ImageNet top-1 (\%)  &Real top-1 (\%)& Imagenet-v2 \\
\hline\hline

\multirow{3}{*}{} 
&ResNet-50 ~\cite{he2016deep}&C& 25 &4.1& 76.2& 82.5& 63.3\\
$224^2$&ResNet-101 ~\cite{he2016deep}&C &45& 7.9& 77.4& 83.7 &65.7\\
&ResNet-152 ~\cite{he2016deep}&C& 60& 11& \textbf{78.3} &\textbf{84.1}& \textbf{67.0}\\\hline

&RedNet-50 ~\cite{li2021involution}&I& 15.5 &-& 78.4& -& -\\
$224^2$&RedNet-101 ~\cite{li2021involution}&I &25.6& -& 79.1& - &-\\
&RedNet-152 ~\cite{li2021involution}&I& 34& -& \textbf{79.3} &-& \\\hline

\multirow{6}{*}{}
&DeiT-S ~\cite{touvron2021training}&T& 22& 4.6& 79.8& 85.7 &68.5\\
&DeiT-B ~\cite{touvron2021training}&T& 86& 17.6 &81.8 &\textbf{86.7} &\textbf{71.5}\\ \cline{2-8}

&PVT-Small ~\cite{wang2021pyramid}&T&25& 3.8& 79.8& – &–\\
&PVT-Medium ~\cite{wang2021pyramid}&T& 44& 6.7& 81.2& – &–\\
&PVT-Large~\cite{wang2021pyramid}&T& 61 &9.8& 81.7& –& –\\\cline{2-8}
&Cross-ViT-S~\cite{chen2021crossViT}&T & 26.7& 5.6 & 81.0 &- &-\\
&Cross-ViT-B~\cite{chen2021crossViT}&T & 104.7& 21.2 & 82.2  &-  &-\\\cline{2-8}

&T2T-ViTt-14 ~\cite{yuan2021tokens}&T &22 &6.1 &80.7& – &–\\
&T2T-ViTt-19 ~\cite{yuan2021tokens}&T& 39& 9.8& 81.4& –& –\\
&T2T-ViTt-24~\cite{yuan2021tokens}&T& 64& 15.0 &82.2& – &–\\\cline{2-8}
$224^2$&TNT-S ~\cite{han2021transformer}&T& 24 &5.2 &81.3& –& –\\
&TNT-B ~\cite{han2021transformer}&T &66& 14.1 &82.8 &– &–\\ \cline{2-8}
&CiT-Ti~\cite{ren2022co}&T& 6 &  -& 75.3 &-& -\\
&CiT-S~\cite{ren2022co}&T& 22 &  -& 82.0 &-& -\\\cline{2-8}
&Visformer~\cite{chen2021visformer}&T &	 40.2 & 4.9 & 82.3&- &-\\

&Swin-S ~\cite{liu2021swin}&T &50 & 8.7& 83.2 &-& -\\
&Swin-B ~\cite{liu2021swin}&T &88 & 15.4& 83.5& - &-\\\cline{2-8}
&LIT-Ti~\cite{pan2022less}&T & 19&  3.6&  81.1 & 86.6 & 70.4\\
&LIT-S~\cite{pan2022less}&T & 27&  4.1&  81.5 &  86.4&  70.4\\
&LIT-M~\cite{pan2022less}&T & 48&  8.6&  83.0&  87.3&  72.0\\
&LIT-B~\cite{pan2022less}&T & 86 & 15.0 & 83.4 & 87.6 & 72.8\\\cline{2-8}

&LITv2-S~\cite{panfast}&T &28 &3.7& 82.0&- &-\\
&LITv2-M~\cite{panfast}&T & 49& 7.5 & 83.3&- &-\\
&LITv2-B ~\cite{panfast}&T & 87 &13.2 & 83.6&- &-\\\cline{2-8}
&Twins-PCPVT-B~\cite{chu2021twins}&T & 43.8 & 6.7 &  82.7 &- &-\\
&Twins-SVT-B~\cite{chu2021twins}&T & 56 & 8.6  & 83.2  &-  &-\\
&Twins-PCPVT-L~\cite{chu2021twins}&T & 60.9 & 9.8 &  83.1 &- &-\\
&Twins-SVT-L~\cite{chu2021twins}&T & 99.2 & 15.1  & 83.7  &-  &-\\
\cline{2-8}

&ViL-Small ~\cite{zhang2021multi}&T & 24.6&  4.9& 82.4& -& -\\
&ViL-Medium ~\cite{zhang2021multi}&T &39.7& 8.7& 83.5 &-& -\\
&ViL-Base ~\cite{zhang2021multi}&T & 55.7&  13.4 &83.7& -& -\\\cline{2-8}

&RegionViT-Ti+~\cite{chen2022regionViT}&T &	14.3&	2.7&	81.5 &- &-\\
&RegionViT-M+~\cite{chen2022regionViT}&T &	42.0&	7.9&	83.4 &- &-\\
&RegionViT-B~\cite{chen2022regionViT}&T &	72.7&	13.0&	83.2 &- &-\\
&RegionViT-B+~\cite{chen2022regionViT}&T &	73.8&	13.6&	83.8 &- &-\\
\cline{2-8}

&DAT-T ~\cite{xia2022vision}&T & 29& 4.6&  82.0&- &-\\
&DAT-S ~\cite{xia2022vision}&T & 50& 9.0&  83.7&- &-\\
&DAT-B  ~\cite{xia2022vision}&T &  88& 15.8&  84.0&- &-\\\cline{2-8}

&SE-CoTNetD-50 ~\cite{li2022contextual}&T &	23.1&	4.1&	81.6 &- &-\\
&SE-CoTNetD-101 ~\cite{li2022contextual}&T &	40.9&	8.5&	83.2 &- &-\\
&SE-CoTNetD-152 ~\cite{li2022contextual}	&T &55.8&	17.0&	84.0 &- &-\\
\cline{2-8}

&ViL-LS-Medium~\cite{zhu2021long}&T & 39.8 & 8.7& 83.8& -& -\\
&ViL-LS-Base~\cite{zhu2021long}&T & 55.8& 13.4& 84.1& -& -\\\cline{2-8}

&UniNet-B1~\cite{liu2022uninet}&CT &  11.5&  1.1& 80.8&- &-\\ \cline{2-8}
&CoaT-Lite-Mini~\cite{xu2021co}&CT & 10& 6.8 & 80.8 &- &-\\
&CoaT-Lite-Small~\cite{xu2021co}&CT & 20& 4.0 & 81.9  &-  &-\\\cline{2-8}
&CvT-13 ~\cite{wu2021CvT}&CT &20 &4.5 &81.6& 86.7& 70.4\\
&CvT-21 ~\cite{wu2021CvT}&CT& 32& 7.1& 82.5& 87.2& \textbf{71.3}\\\cline{2-8}
&LV-ViT-$S^*$~\cite{jiang2021all}&CT& 26&    6.6B& 83.3& 88.1& -\\
&LV-ViT-$M^*$~\cite{jiang2021all}&CT& 56 &  16.0 &84.1 &\textbf{88.4}& -\\\cline{2-8}

\hline
       
\end{tabular}
\caption{Performance of the various models on ImageNet1K \protect\cite{deng2009imagenet}, ImageNet Real \protect\cite{beyer2020we} and ImageNet V2 matched frequency \protect\cite{recht2019imagenet} with image size $224^2$ .  We report these numbers from  CvT \protect\cite{wu2021CvT} paper. C  stands for Convolution, I stands for Involution, T tends for Transformer, and CT stands for Convolution  Transformers. $*$ means extra data used while training the model.}\label{EVT_tab_1}
\end{table*}

\begin{table*}[tb]
\centering
\begin{tabular}{p{3em}|p{17em}p{3em}p{3em}p{3em}p{4em}p{4em}p{2em}}
\hline
Image Size & Method & Network Type  & \#Param (M) & FLOPS (G) & ImageNet top-1 (\%)  &Real top-1 (\%)& Imagenet-v2 \\
\hline\hline


&CSwin-T ~\cite{dong2022cswin}&T & 23 & 4.3 &  82.7&- &-\\
&CSwin-S ~\cite{dong2022cswin}&T & 35 & 6.9  & 83.6&- &-\\
&CSwin-B ~\cite{dong2022cswin}&T & 78  &15.0  & 84.2&- &-\\\cline{2-8}

&DaViT-Ti  ~\cite{ding2022daViT}&T & 28.3 & 4.5 &  82.8&- &-\\
&DaViT-S ~\cite{ding2022daViT}&T & 49.7 & 8.8 & 84.2&- &-\\
&DaViT-B~\cite{ding2022daViT}&T & 87.9 & 15.5  &84.6&- &-\\\cline{2-8}

&CaiT-M36~\cite{touvron2021going}&T& 271&  53.7&  85.1& \textbf{89.3}& -\\\cline{2-8}

&MViTv2-T~\cite{li2022mvitv2}&T &24& 4.7& 82.3&- &-\\
&MViTv2-S~\cite{li2022mvitv2}&T &35& 7& 83.6&- &-\\
&MViTv2-B~\cite{li2022mvitv2}&T &52& 10.2& 84.4&- &-\\
&MViTv2-L~\cite{li2022mvitv2}&T &218& 42.1& 85.3&- &-\\\cline{2-8}

&Wave-ViT-$S^*$~\cite{yao2022wave}&T &22.7&	4.7	&83.9 &- &-\\
&Wave-ViT-$B^*$~\cite{yao2022wave}&T &33.5&	7.2	&84.8 &- &-\\
&Wave-ViT-$L^*$~\cite{yao2022wave}&T &57.5&	14.8&	\textbf{85.5}  &- &-\\\cline{2-8}

&CMT-Ti ~\cite{guo2022cmt}&CT&9.5&0.6& 79.1&-& -\\
&CMT-XS ~\cite{guo2022cmt}&CT&15.2&1.5& 81.8&-& -\\
&CMT-S~\cite{guo2022cmt}&CT& 25.1 &  4& 83.5 &-& -\\
&CMT-B ~\cite{guo2022cmt}&CT&45.7&9.3& 84.5&-& -\\
&CMT-L ~\cite{guo2022cmt}&CT&74.7&19.5& 84.8&-& -\\\cline{2-8}

&MaxViT-T~\cite{tu2022maxvit}&CT & 31&  5.6&  83.62&- &-\\
&MaxViT-S~\cite{tu2022maxvit}&CT & 69& 11.7& 84.45&- &-\\
&MaxViT-B ~\cite{tu2022maxvit}&CT & 120& 23.4&84.95&- &-\\
&MaxViT-L~\cite{tu2022maxvit}&CT &212& 43.9& 85.17&- &-\\\cline{2-8}

&UniFormer-S~\cite{li2022uniformer}&CT & 22 & 3.6 & 82.9&- &-\\
&UniFormer-B ~\cite{li2022uniformer}&CT & 50 & 8.3  &83.9&- &-\\
&UniFormer-L~\cite{li2022uniformer}&CT & 100 & 12.6  & \textbf{85.6}&- &-\\

\hline
       
\end{tabular}
\caption{This is an extension of table-~\ref{EVT_tab_1}, which shows performance of the various models on ImageNet1K \protect\cite{deng2009imagenet}, ImageNet Real \protect\cite{beyer2020we} and ImageNet V2 matched frequency \protect\cite{recht2019imagenet} with image size $224^2$ .  We report these numbers from  CvT \protect\cite{wu2021CvT} paper. C  stands for Convolution, I stands for Involution, T tends for Transformer, and CT stands for Convolution  Transformers. $*$ means extra data used while training the model.}\label{EVT_tab_2_extention}
\vspace{-1.2em}
\end{table*}
\begin{table*}[tb]
\centering
\begin{tabular}{p{3em}|p{17em}p{3em}p{3em}p{3em}p{4em}p{4em}p{2em}}
\hline
Image Size & Method & Network Type  & \#Param (M) & FLOPS (G) & ImageNet top-1 (\%)  &Real top-1 (\%)& Imagenet-v2 \\
\hline\hline

&Mixer-B/16 ~\cite{tolstikhin2021mlp}&M & 59& 12.7& 76.4&- &-\\ \cline{2-8}

&gMLP-S~\cite{liu2021pay}&M & 20& 4.5& 79.6&- &-\\ 
&gMLP-B~\cite{liu2021pay}&M & 73& 15.8&81.6&- &-\\ \cline{2-8}
&ResMLP-S12~\cite{touvron2022resmlp}&M &15& 3.0&76.6&- &-\\
&ResMLP-S24~\cite{touvron2022resmlp}&M &30& 6.0& 79.4&- &-\\
&ResMLP-B24~\cite{touvron2022resmlp}&M &116& 23.0& 81.0&- &-\\\cline{2-8}

&$S^2$-MLP-wide~\cite{yu2022s2}&M & 71& 14.0& 80.0&- &-\\
&$S^2$-MLP-deep~\cite{yu2022s2}&M & 51& 10.5& 80.7&- &-\\\cline{2-8}
&ViP-Small/7 ~\cite{hou2022vision}&M & 25& 6.9& 81.5&- &-\\
&ViP-Medium/7 ~\cite{hou2022vision}&M & 55& 16.3& 82.7&- &-\\
&ViP-Large/7 ~\cite{hou2022vision}&M & 88& 24.4& 83.2&- &-\\\cline{2-8}

&CycleMLP-B1~\cite{chencyclemlp}&M & 15&2.1& 78.9&- &-\\
&CycleMLP-B2~\cite{chencyclemlp}&M &27& 3.9& 81.6&- &-\\
&CycleMLP-B3~\cite{chencyclemlp}&M & 38& 6.9& 82.4&- &-\\
&CycleMLP-B4~\cite{chencyclemlp}&M &52& 10.1& 83.0&- &-\\
&CycleMLP-B5~\cite{chencyclemlp}&M & 76& 12.3& 83.2&- &-\\\cline{2-8}
&AS-MLP-T  ~\cite{lianmlp}&M &28& 4.4& 81.3&- &-\\
&AS-MLP-S  ~\cite{lianmlp}&M & 50& 8.5& 83.1&- &-\\
&AS-MLP-B  ~\cite{lianmlp}&M & 88& 15.2& 83.3&- &-\\\cline{2-8}

&Wave-MLP-T~\cite{tang2022image}&M & 1&2.4& 80.6&- &-\\
$224^2$&Wave-MLP-S~\cite{tang2022image}&M & 30& 4.5& 82.6&- &-\\
&Wave-MLP-M~\cite{tang2022image}&M & 44& 7.9& 83.4&- &-\\
&Wave-MLP-B~\cite{tang2022image}&M & 63& 10.2& 83.6&- &-\\\cline{2-8}
&Hire-MLP-Tiny ~\cite{Guo_2022_CVPR}&M & 18& 2.1& 79.7&- &-\\
&Hire-MLP-Small~\cite{Guo_2022_CVPR}&M & 33& 4.2& 82.1&- &-\\
&Hire-MLP-Base~\cite{Guo_2022_CVPR}&M & 58& 8.1& 83.2&- &-\\
&Hire-MLP-Large~\cite{Guo_2022_CVPR}&M & 96& 13.4& 83.8&- &-\\\cline{2-8}

&DynaMixer-S~\cite{wang2022dynamixer}&M & 26& 7.3& 82.7&- &-\\
&DynaMixer-M~\cite{wang2022dynamixer}&M & 57& 17.0& 83.7&- &-\\
&DynaMixer-L~\cite{wang2022dynamixer}&M & 97& 27.4& \textbf{84.3}&- &-\\\cline{2-8}
&PoolFormer-S12~\cite{yu2022metaformer}&P &  12&  1.9&  77.2&- &-\\ 
&PoolFormer-S24~\cite{yu2022metaformer}&P &21&  3.5&  80.3&- &-\\
&PoolFormer-S36 ~\cite{yu2022metaformer}&P & 31&  5.1&  81.4&- &-\\
&PoolFormer-M36 ~\cite{yu2022metaformer}&P & 56&  9.0&  82.1&- &-\\
&PoolFormer-M48 ~\cite{yu2022metaformer}&P & 73&  11.8&  \textbf{82.5}&- &-\\\cline{2-8}
\hline
      
\end{tabular}
\vspace{-0.7em}
\caption{This table shows the performance of the various MLP-like Transformer models on ImageNet1K \protect\cite{deng2009imagenet}, ImageNet Real \protect\cite{beyer2020we} and ImageNet V2 matched frequency \protect\cite{recht2019imagenet} with image size $224^2$ .  We report these numbers from  CvT \protect\cite{wu2021CvT} paper.  M stands for MLP-Mixer, and  P stands for Pooling Network.  $*$ means extra data used while training the model.}\label{EVT_tab_1_extention}
\vspace{-1em}
\end{table*}

MLP-mixer~\cite{tolstikhin2021mlp} model use multi-layer perceptions (MLPs) to mix input token without using CNNs or self-attention network. It uses two types of layers one MLP layer to combine the location feature of image patches and the other one for mixing spatial information. The MLP-mixer models achieve competitive scores on image classification benchmarks. Lian et al. proposed axial shifted MLP (AS-MLP)  ~\cite{lianmlp} network for vision tasks. AS-MLP pays more attention to local feature interaction by using axial shifted channels of the feature maps. It captures local dependency by capturing the flow of the information in axial directions, as in CNNs.  
MLP-mixer~\cite{tolstikhin2021mlp} and ResMLP~\cite{touvron2022resmlp} flattened the input image patches and fed them into the transformer encoder to mix the input tokens linearly using MLP networks. It is hard to capture the spatial information in the image. To avoid this issue, Guo et al. have proposed a Hire-MLP~\cite{Guo_2022_CVPR} network which contains  MLP architecture
via Hierarchical rearrangements in two levels. The inner region rearrangement captures local information inside the spatial regions, and cross-region rearrangement enables the communication between two regions. It captures global context by circularly shifting all tokens along the spatial direction. MLP-mixer~\cite{tolstikhin2021mlp}, ResMLP~\cite{touvron2022resmlp} and gMLP~\cite{liu2021pay} architecture highly depend on the image size. These MLP models are in quadratic O($N^2$) computational complexity in nature, which are infeasible for object detection and semantic segmentation task. To solve this issue, Chen et al. ~\cite{chencyclemlp} has proposed a CycleMLP network, which mainly it can cope with variable image size and provides linear complexity with image size using local windows. Recently MLP based architecture based on fully connected layers of archives competing for performance like CNN and transformers. We split images as multiple tokens(patches) and directly aggregated them with fixed weights by neglecting varying semantic information of tokens from different images. To handle this issue, Tang et al. have proposed Wave-MLP~\cite{tang2022image} method in which each token has two parts, amplitude, and phase part. The amplitude is the original feature, and the phase terms are a complex value chaining according to the image's semantic content by modulating the relation between tokens and fixed weights.

Recently the attention-based module of the transformer is replaced by the spatial MLPs, as shown in the figure-\ref{fig_trans}. These MLP-based models are performing well compared to attention-based models. Generally, a transformer architecture-specific token mixing module is necessary for the model's performance. Yu et al. have proposed simple methods to replace the attention module of the transformer with a simple spatial pooling operator to conduct only basic token mixing, and this method is known as PoolFormer.

\begin{table*}[!htb]
\centering
\begin{tabular}{p{3em}|p{16em}p{3em}p{3em}p{3em}p{4em}p{4em}p{2em}}
\hline
Image Size & Method & Network Type  & \#Param (M) & FLOPS (G) & ImageNet top-1 (\%)  &Real top-1 (\%)& Imagenet-v2 \\
\hline\hline

\multirow{3}{*}{} 
$600^2$&EfficientNet-B7~\cite{tan2019efficientnet}&C& 43& 19& \textbf{84.3} &-& -\\\hline

$256^2$&BoTNet--S1-128~\cite{srinivas2021bottleneck}&T& 79.1  & 19.3& 84.2& -& -\\

&UniNet-B2 ~\cite{liu2022uninet}&CT &  16.2& 2.2 &82.5&- &-\\

&CMT-B~\cite{guo2022cmt}&CT& 45.7 &  9.3&\textbf{ 84.5} &-& -\\ \hline

&CMT-L~\cite{guo2022cmt}&CT& 74.7 &  19.5& 84.8 &-& -\\
&UniNet-B3~\cite{liu2022uninet}&CT &  24& 4.3 &83.5&- &-\\

$288^2$&LV-ViT-$L^*$~\cite{jiang2021all}&CT& 150 & 59.0& \textbf{85.3}& \textbf{89.3}& -\\\hline

\multirow{2}{*}{}
&ViT-B/16 ~\cite{dosoViTskiy2020image}&T& 86 & 55.5& 77.9& 83.6& –\\
&ViT-L/16 ~\cite{dosoViTskiy2020image}&T&307 &  191.1& 76.5& 82.2& –\\\cline{2-8}

&DeiT-B ~\cite{touvron2021training}&T&86&  55.5& 83.1& -& -\\
&Swin-B ~\cite{liu2021swin}&T & 88 & 47.1 &84.5& -& -\\
&ViL-LS-Medium ~\cite{zhu2021long}&T &39.9&  28.7& 84.4& -& -\\ 
&LITv2-B~\cite{panfast}&T  & 87 &39.7&  84.7&- &-\\
&DAT-B  ~\cite{xia2022vision}&T  & 88& 49.8& 84.8&- &-\\

$384^2$&BoTNet-S1-128~\cite{srinivas2021bottleneck}&T& 79.1&   45.8& 84.7& -& -\\
&CaiT-S36~\cite{touvron2021going}&T& 68 &  48.0& \textbf{85.4} &\textbf{89.8}& -\\\cline{2-8}
&CSwin-T  ~\cite{dong2022cswin}&T & 23 & 14.0 &  84.3&- &-\\
&CSwin-S  ~\cite{dong2022cswin}&T & 35 & 22.0 & 85.0&- &-\\
&CSwin-B  ~\cite{dong2022cswin}&T & 78 & 47.0  &\textbf{85.4}&- &-\\\cline{2-8}

&CvT-13~\cite{wu2021CvT}&CT& 20 &16.3& 83.0&87.9& 71.9\\
&CvT-21~\cite{wu2021CvT}&CT &32& 24.9& 83.3& 87.7& \textbf{71.9}\\\cline{2-8}
&UniNet-B5 ~\cite{liu2022uninet}&CT &   72.9&  20.4 &84.9&- &-\\\cline{2-8}
&LV-ViT-$S^*$~\cite{jiang2021all}&CT& 26&   22.2& 84.4& 88.9& -\\
&LV-ViT-$M^*$~\cite{jiang2021all}&CT& 56  &42.2& 85.4& \textbf{89.5}& -\\ \cline{2-8}

&UniFormer-S~\cite{li2022uniformer}&CT & 22 & 11.9 & 84.6&- &-\\
&UniFormer-B~\cite{li2022uniformer}&CT  & 50 & 27.2 &86.0&- &-\\
&UniFormer-L~\cite{li2022uniformer}&CT & 100  &39.2  & 86.3&- &-\\\cline{2-8}
&MaxViT-S~\cite{tu2022maxvit}&CT &  69& 36.1& 85.74&- &-\\
&MaxViT-B~\cite{tu2022maxvit}&CT &  120& 74.2& 86.34&- &-\\
&MaxViT-L ~\cite{tu2022maxvit}&CT & 212& 133.1& \textbf{86.40}&- &-\\\cline{2-8}

\hline
\multirow{3}{*}{}

$448^2$& CaiT-M36~\cite{touvron2021going} & T & 271 &  247.8&\textbf{86.3}& \textbf{90.2} & -\\\cline{2-8}
&UniNet-B6 ~\cite{liu2022uninet}&CT &  117& 51 &85.6&- &-\\
&LV-ViT-$L^*$~\cite{jiang2021all}&CT& 150 & 157.2& \textbf{85.9}& 89.7& -\\ 
\hline

&LV-ViT-$L^*$~\cite{jiang2021all}&CT& 151 & 214.8& 86.4& \textbf{90.1}& -\\ 
&MaxViT-T~\cite{tu2022maxvit}&CT &  31& 33.7& 85.72&- &-\\
$512^2$&MaxViT-S ~\cite{tu2022maxvit}&CT & 69& 67.6& 86.19&- &-\\
&MaxViT-B~\cite{tu2022maxvit}&CT &  120& 138.5& 86.66&- &-\\
&MaxViT-L~\cite{tu2022maxvit}&CT &  212&245.4& \textbf{86.70}&- &-\\
\hline
       
\end{tabular}
\caption{Performance of the various models on ImageNet1K \protect\cite{deng2009imagenet}, ImageNet Real \protect\cite{beyer2020we} and ImageNet V2 matched frequency \protect\cite{recht2019imagenet} for various image sizes. We report these numbers from  CvT \protect\cite{wu2021CvT} paper.  C  stands for Convolution, I stands for Involution, T stands for Transformer, and CT stands for Convolution  Transformers. $*$ means extra data used while training the model.
}
\vspace{-1.5em}
\label{EVT_tab_1_1}
\end{table*}

 TopFormer: As we know, the computational cost of a vision transformer is very high for dense prediction tasks like semantic segmentation tasks on mobile networks. 
 Zhang et al. ~\cite{zhang2022topformer} have proposed a mobile-friendly architecture such as a Token Pyramid vision transformer (TopFormer) for the dense prediction task. The method consists of a Token pyramid module semantic extractor, injection module, and segmentation head on it. The token pyramid module takes an image as an image put and produces a token pyramid, which is fed to the semantic extractor to produce scale-aware semantics, which is injected into a token of the corresponding scale for augmenting representation using the injection module. Finally, The augmented token pyramid to perform the segmentation task using a segmentation head.

Dong et al. have proposed a Cross-Shaped Window transformer model CSWin~\cite{dong2022cswin} to compute self-attention in horizontal and vertical strips in parallel to form cross-shape windows. The main difference from vanilla vision transformers are  1.) CSWin transformer replaces multi-headed self-attention with Cross Shaped Window Self Attention. 2.) it introduces local inductive bias using  Locally-enhanced Positional Encoding (LePE), which is added parallel to the self-attention module. The CSWin transformer is efficient in terms of accuracy and good representation feature for downstream tasks.

RegionViT~\cite{chen2022regionViT} has proposed a new architecture that adapts a pyramid structure to capture regional-to-local attention rather than global attention for the vision transformer. The method first generates regional tokens and local tokens from images of different patch sizes. Regional self-attention captures global information among all regional tokens, and local self-attention exchanges the information among one regional token and associates with the local token via self-attention.

\begin{table*}[htb]
\centering
\begin{tabular}{p{3em}|p{16em}p{3em}p{3em}p{3em}p{4em}p{4em}p{2em}}
\hline
Image Size & Method & Network Type  & \#Param (M) & FLOPS (G) & ImageNet top-1 (\%)  &Real top-1 (\%)& Imagenet-v2 \\
\hline\hline

\multirow{3}{*}{} 
$480^2$&BiT-M~\cite{kolesnikov2020big} &T& 928&837& 85.4& –& –\\ \hline

&ViT-B/16~\cite{dosoViTskiy2020image}&T &86& 55.5& 84.0 &88.4& –\\
&ViT-L/16~\cite{dosoViTskiy2020image}&T& 307&191.1& 85.2& 88.4& –\\
$384^2$&ViT-H/16~\cite{dosoViTskiy2020image}&T& 632&– &85.1 &88.7 &–\\ \cline{2-8}
 &CvT-13~\cite{wu2021CvT}&CT& 20&16 &83.3& 88.7 &72.9\\
&CvT-21~\cite{wu2021CvT}&CT& 32 &25& 84.9& 89.8& 75.6\\
 &CvT-W24~\cite{wu2021CvT}&CT& 277&193.2& 87.7& 90.6 &78\\
 \hline
      
\end{tabular}
\caption{Performance of the various transformer models pre-trained on ImageNet21K \protect\cite{deng2009imagenet} and fine-tuned on ImageNet21K \protect\cite{deng2009imagenet} ImageNet Real \protect\cite{beyer2020we} and ImageNet V2 matched frequency \protect\cite{recht2019imagenet} for image size $384 \times 384$, except BiT-M \protect\cite{kolesnikov2020big}, which is fine-tuned on image size $480 \times 480$. We report these numbers from  CvT\protect\cite{wu2021CvT} paper. T tends for Transformer, CT tends for Convolution  Transformers.}

\label{EVT_tab_1_2}
\end{table*}

\begin{table*}[htb]
\centering
\begin{tabular}{p{2em}|p{13em}p{3em}p{3em}p{4em}p{4em}p{3em}}
\hline
Image Size & Method  & \#Param (M) & FLOPS (G) & ImageNet top-1 (\%)  & Attention Network  & Extra Label \\
\hline\hline

\multirow{3}{*}{} 
&Fnet~\cite{lee2021fnet} &15& 2.9& 71.2 &\textcolor{red}{\xmark}&\textcolor{red}{\xmark}\\ \cline{2-7}
&GFNet-Ti~\cite{rao2021global} &7& 1.3& 74.6 &\textcolor{red}{\xmark}&\textcolor{red}{\xmark}\\ 
$224^2$&GFNet-XS~\cite{rao2021global} &16& 2.9& 78.6&\textcolor{red}{\xmark}&\textcolor{red}{\xmark}\\
&GFNet-S~\cite{rao2021global} &25& 4.5& 80.0&\textcolor{red}{\xmark}&\textcolor{red}{\xmark}\\
&GFNet-B~\cite{rao2021global} &43 & 7.9 & 80.7&\textcolor{red}{\xmark}&\textcolor{red}{\xmark}\\\cline{2-7}
&AFNO-S/4~\cite{guibas2021efficient} &16& 15.3& \textbf{80.9}&\textcolor{red}
{\xmark}&\textcolor{red}{\xmark}\\\cline{2-7}

&Wave-ViT-S~\cite{yao2022wave} &19.8& 4.3& 82.7&\textcolor{green}{\cmark}&\textcolor{red}{\xmark}\\
&Wave-ViT-S$^*$~\cite{yao2022wave}&22.7& 4.7& 83.9&\textcolor{green}{\cmark}&\textcolor{green}{\cmark}\\
&Wave-ViT-B$^*$~\cite{yao2022wave}&33.5& 7.2& 84.8 &\textcolor{green}{\cmark}&\textcolor{green}{\cmark}\\
&Wave-ViT-L$^*$ ~\cite{yao2022wave}&57.5& 14.8& \textbf{85.5}&\textcolor{green}{\cmark}&\textcolor{green}{\cmark}\\
\hline
&GFNet-XS~\cite{rao2021global} &18& 8.4& 80.6&\textcolor{red}{\xmark}&\textcolor{red}{\xmark}\\
$384^2$&GFNet-S~\cite{rao2021global} &28& 13.2& 81.7&\textcolor{red}{\xmark}&\textcolor{red}{\xmark}\\
&GFNet-B~\cite{rao2021global} &47 &23.3 & \textbf{82.1} &\textcolor{red}{\xmark}&\textcolor{red}{\xmark}\\
\hline
      
\end{tabular}
\caption{This table shows the performance analysis of spectral transformer models trained on ImageNet1K\protect\cite{deng2009imagenet} for image size $224 \times 224$ and $384 \times 384$. It compares the number of parameters, number of FLOPs, and Top-1 accuracy of various spectral vision transformer models. The red tick mark in the attention column indicates that the attention network is not used in the transformer model. $*$ indicates that WaveViT\protect\cite{yao2022wave} uses extra training data while training.}
\vspace{-0.7em}
\label{EVT_tab_5}
\end{table*}

 UniFormer~\cite{li2022uniformer} model is a more generalizing model considering both Convolution and Self-attention for the Visual Recognition task. As CNNs based models are more efficient for decreasing local redundancy by convolution within a small
neighborhood, the limited receptive field makes it hard to capture global dependency. Whereas transformer-based models can effectively capture
long-range dependency via self-attention while finding similarity comparisons among all the tokens lead to high redundancy. The UniFormer model uses  Multi-Head Relation Aggregator(MHRA)  blocks, which are modeled with local and global tokens affinity in the shallow and deep layer, respectively,  to reduce redundancy and provide an efficient representing feature. The main difference in the UniFormer model compare with the transformer is it uses MHRA, while the transformer uses MSA (Multi-headed Self Attention). 

The scalability of self-attention is restricted to the image size. That is, the computation complexity increase as image size increases. To avoid this issue, Tu et al. proposed the MaxVit~\cite{tu2022maxvit} method to efficiently scale self-attention. The technique contains two concepts, one block local and dilated global attention. The model provides local-global spatial interaction on various input image resolutions with linear complexity. MViTv2~\cite{li2022mvitv2} has improved the performance of  Multi-scale Vision Transformers for Classification and Detection tasks. The  MViTv2 method decomposed location distance to inject the position information to the transformer blocks and residual pooling connections to composite the effect of pooling in the attention computation.

Pan et al. have proposed an efficient method (LIT)~\cite{pan2022less}, which pays less attention to the self-attention module in the vision Transformer. The early self-attention layers focus on local patterns and provide minor benefits in hierarchal vision transformers. Pan et al. further improve the LIT model with hilo attention~\cite{panfast} method to make a fast vision transformer. Venkataramanan et al.~\cite{venkataramanan2023skip} have proposed a skip-attention method to improve vision transformers by paying less attention.  Similarly, GC-ViT~\cite{hatamizadeh2022global} and Beit~\cite{bao2022beit} proposed various attention mechanisms to improve the performance of the vision transformer on ImageNet-1K\cite{deng2009imagenet} dataset.

We compare top-1 accuracy (\%) of various transformer models for image size $224^2$ on the ImageNet dataset over model parameter in millions(M) as shown in the figure-\ref{fig4}.

\begin{table*}[htb]
\centering
\begin{tabular}{p{15em}p{3em}p{3em}p{3em}p{3em}p{3em}p{3em}p{3em}}
\hline
Method & \#FLOPS (G) &\#Param (M) & CIFAR 10 & CIFAR 100  & Pet & Flower & Cars \\
\hline\hline

ViT-B/16~\cite{dosoViTskiy2020image} & 55.4& 86 &98.1 &87.1&- &89.5& -\\
ViT-L/16~\cite{dosoViTskiy2020image} & 190.7 &307 &97.9& 86.4&-& 89.7& -\\
DeiT-B ~\cite{touvron2021training}&17.6 & 85.8  & 99.1 & \textbf{90.8} & -& 98.4&92.1\\
TNT-S↑384~\cite{han2021transformer}&17.3 & 23.8& 98.7&90.1& 94.7& \textbf{98.8}&-\\
CaiT-S↑384~\cite{touvron2021going}&12.9  & 24.2 & 99.1 & \textbf{90.8} &   94.9 &98.6&94.1\\
GFNet-XS\cite{rao2021global} &2.9& 16& 98.6& 89.1& -&98.1 &92.8\\
GFNet-H-B\cite{rao2021global}& 8.6& 54& 99.0& 90.3&  -&\textbf{98.8} &93.2\\
CMT-S~\cite{guo2022cmt}&4.04 &25.1&\textbf{99.2}& 91.7&95.2& 98.7 &\textbf{94.4}\\
RegionViT-S~\cite{chen2022regionViT}&- & -& 98.9& 90.0& 95.3&-& 92.8 \\
RegionViT-M~\cite{chen2022regionViT}& -& -&99.0& \textbf{90.8}& \textbf{95.5}&- &91.9\\
 \hline

BiT-M~\cite{kolesnikov2020big}&-& 928& 98.91& 92.17 &94.46& 99.30 &- \\
ViT-B/16~\cite{dosoViTskiy2020image}&- &86& 98.95& 91.67 &94.43& 99.38&- \\
ViT-L/16~\cite{dosoViTskiy2020image}& -& 307 &99.16& 93.44& 94.73& 99.61& -\\
ViT-H/16~\cite{dosoViTskiy2020image}& -& 632 & 99.27 &93.82 &\textbf{94.82} & 99.51 &- \\
CvT-13~\cite{wu2021CvT} &-& 20& 98.83& 91.11& 93.25 &99.50&- \\
CvT-21~\cite{wu2021CvT} &-& 32 & 99.16 & 92.88& 94.03 &99.62&-\\
CvT-W24~\cite{wu2021CvT}&- & 277 & \textbf{99.39}  &\textbf{94.09}  &94.73 &\textbf{99.72} &  \\


\hline
\end{tabular}
\vspace{-0.5em}
\caption{ Transfer Learning performance on CIFAR10 \protect\cite{krizhevsky2009learning}, CIFAR100 \protect\cite{krizhevsky2009learning}, Pet \protect\cite{parkhi2012cats}, Flower \protect\cite{nilsback2008automated} and Cars \protect\cite{krause20133d} dataset. We reported top-1 accuracy, Number of FLOPS, and parameters for various transformer models on these datasets. The top block of the table indicates, models are  pre-trained on ImageNet 1k \protect\cite{deng2009imagenet} and the bottom block indicates,  the models are pre-trained on ImageNet 22k\protect\cite{deng2009imagenet}. }\label{EVT_tab_2}
\vspace{-1em}
\end{table*}
\subsection{Spectral complexity} 

Efficient transformers can be designed to speed up transformer encoder architecture by replacing the self-attention network with linear transformations that mix input tokens. The self-attention layer of the transformer is replaced by a parameterized Fourier transformation (Fnet)~\cite{lee2021fnet}, which is then followed by a non-linearity and feed-forward network. Compared to BERT, this network is 80 percent faster and can archive 92 to 97 percent of the transformer performance. The Global Frequency network (GFnet)~\cite{rao2021global} proposes a depth-wise global convolution for token mixing. GFnet involves three steps: Spatial token mixing via Fast Fourier Transform (FFT), frequency gating, and inverse FFT for token demixing. GFnet is not involved in channel mixing, is expensive for higher solution images as sequence length increases, and is not adaptive. Guibias et al. ~\cite{guibas2021efficient} formulated the token mixing task as an operator-learning task that learns mapping among continuous functions in infinite dimensional space. Li et al. ~\cite{li2020fourier} discuss solving Partial Differential Equations (PDE) using a Fourier Neural Operator (FNO).
FNO works well in continuous domains. Adapting FNO for a vision domain with high-resolution image inputs requires modification in the design architecture of FNO from PDE. This is because high images have discontinuities due to edges and other structures. Additionally, the channel mixing FNO depends on the channel size, which has quadratic complexity. The block-diagonal structure is used on channel mixing weight to handle this channel mixing issue. The author shared weights across the tokens of MLP layers for parameter efficiency and introduced sparsity in the frequency domain using soft thresholding for generalization. These solutions combine, known as  Adaptive Fourier neural Operator (AFNO).   In Wave-ViT~\cite{yao2022wave}, the author has discussed the quadratic complexity of the self-attention network of the transformer model with input patch numbers. People have used downsampling operations using global average polling (GAP) over key/values to solve this issue in the past. It is observed that downsampling operations such as (GAP) are non-invertible,  which causes losing high frequency, such as texture details of the objects. The author has proposed a wavelet vision transformer to perform lossless downsampling using wavelet transform over keys and values. The model performs state-of-the-art results on image recognition, object detection, and instance segmentation tasks. The model is efficient in terms of the number of FLOPS and accuracy. Compared to GFNet~\cite{rao2021global} and AFNO~\cite{guibas2021efficient}, WaveVit~\cite{yao2022wave} uses an attention network with extra labels for training. We compare all sorts of spectral networks in table-~\ref{EVT_tab_5}.
  Another recent work in the spectral domain is Fourier Image transformer (FIT)\cite{buchholz2022fourier}, which uses Fourier domain encoding for image completion tasks, predicting high-resolution output given low-resolution input. This method is demonstrated in computer tomography (CT) image reconstruction tasks.

\subsection{Bias and Fairness Features}
This section discusses inductive bias and how efficiently we can train our transformer models. We start with Inductive Bias (IB) and introduce ViTAE~\cite{xu2021ViTae} work and its variants. ViTAEv2~\cite{zhang2023ViTaev2} lack intrinsic inductive bias (IB) in modeling local visual structure. The model requires massive training data and time to learn inductive bias implicitly. The author has proposed ViTAEv2 method based on the ViT transformer model to explore Inductive intrinsic bias from convolutions. In this method, the author uses spatial reduction modules to downsample and feed the input images into tokens using multiple convolutions with different dilation rates. This model is based on ViTAE~\cite{xu2021ViTae}. Edaelman et al.~\cite{edelman2022inductive}
 has discussed the theoretical analysis of the Inductive biases of the self-attention module. The author shows that the bounded-norm transformer model creates sparse variables, and the single multi-head attention can represent the sparse function of the input sequence. Ren et al.~\cite{ren2022co} have discussed on induction bias of the vision transformer, which is not performing well with insufficient data. The author has introduced a knowledge distillation (KD) based method to help the training of the transformer. The existing methods in KD use heavy convolution neural network-based teacher modules, but in this work, the author uses lightweight modules with different architectural inductive biases as teacher modules (such as CNN-based teacher module and Involution-based teacher module) to co-advise the student transformer model. The author claims that Co-advise based transformer model is more efficient in training and provides better performances compared to existing methods.

















\begin{table*}[]
\centering
\begin{tabular}{p{7em}p{3em}p{3em}p{3em}p{3em}p{3em}p{3em}p{3em}}
\hline
Model & mCE(\%) &   mFR(\%)&  mT5D(\%) &  cAcc(\%)  &  Top-1 (\%)  &  Top-1 (\%)  &  Top-1 (\%)\\
\hline\hline

ResNet-50 & 76.70 &58.00& 82.00 & \textbf{22.30} &25.25& 2.21& 16.76\\
BiT m-r101x3  &58.27 &49.99 &76.71& 03.78 &27.25 &6.41 &28.19\\
ViT L-16 &\textbf{45.45}& \textbf{33.06}& \textbf{50.15 }&20.02& \textbf{40.58} & \textbf{28.10} &\textbf{73.73}\\
\hline
\end{tabular}
\caption{ \protect\cite{paul2022vision}mCEs of different models and methods on ImageNet-C (lower is better). MFRs and mT5Ds on ImageNet-P dataset (lower is better). cAcc tends to challenge accuracy. The cAcc column shows performance on detecting vulnerable image foregrounds from the ImageNet-9 dataset. Columns 6,7, and 8 show top-1 accuracy scores (as percentages) on ImageNet-R, A, and O datasets respectively. }
\label{tab:my_label}
\vspace{-0.7em}
\end{table*} 

\begin{table}
\centering
\begin{tabular}{|l|c|c|c|}
\hline
Method& ImageNet& ImageNet-C& mCE (↓)  \\
\hline\hline
ViT-B/16& 81.43& 58.85& 51.98\\
ViT-L/16& 82.89& 64.11& 45.46\\
Mixer-B/16& 76.47& 47.00& 67.35\\
Mixer-L/16& 71.77& 40.47& 75.84\\
RN18& 69.76& 32.92& 84.67\\
RN50& 76.13& 39.17& 76.70\\
\hline
\end{tabular}

\caption{ This table reports top-1 scores and mCE score of various models on ImageNet \protect\cite{deng2009imagenet}, ImageNet-C \protect\cite{hendrycks2018benchmarking} dataset.}\label{tab_robust_2}
\vspace{-0.7em}
\end{table}

\subsection{Transparency}
Transparency in the Transformer model indicates a clear understanding of the transformer model. Basically, we need to know the details of this model like, what is the transformer model?, How do we train a transformer model? How do we make inferences from those models? How can we explain the inference of the model? We would also like to know more details about the model, such as multi-headed self-attention and positional encoding do? In order to understand the transformer model, we need to know the details. Finally, this will help us to design a more efficient transparent transformer model. Few recent research has been carried out in the field of transparency in the model architecture and training processing.
Park et al.~\cite{park2021vision} have analyzed the vision transformer model and explained "How Do Vision Transformers Work?" with certain examples. The author has analyzed the loss surface for the vision transformers' multi-headed self-attention (MSA). They have also analyzed the behavior of convolution neural network (CNN) and MSA, which seems to be opposite to each other, for example, MSAs are low pass filter and CNNs are high pass filters. This provides transparency while modeling multi-headed self-attention. Raghu et al.~\cite{raghu2021vision} have analyzed "how are vision transformers  solving image classification like convolution networks?"  They study the structure of transformers and CNN and found the ViT are uniform representations across all the layers. The role of self-attention enables early aggregation of global information, and the role of residual connection in ViT propagates features from lower to higher layers. We also consider efficiency in terms of training time and convergence of the model in terms of loss surface. LV-ViT~\cite{jiang2021all} has proposed a different training objective for the vision transformers that compute classification loss with additional trainable class tokens. Location-specific, each patch token is generated by a machine using NFNet ~\cite{brock2021high} image recognition model. The method reformulates the image classification problem into a multi-token-level recognition problem by assigning each patch token with machine-generated supervised location-specific tokens. The method considers all image patch tokens to compute training loss in a dense manner using machine-supervised token labeling. DeiT~\cite{touvron2021training} and AugReg ~\cite{steiner2022train}  have discussed data augmentation methods and regularization methods to train Vision Transformers more efficiently.

\subsection{Robustness} 

Robustness in the transformer is studied in terms of perturbation, common corruption, distributional shift, and natural adversarial examples. Shao et al.~\cite{shao2021adversarial} analyzed the robustness of the transformer model using adversarial perturbation. The authors experimented with a white box and a transformer attack setting. They observe that ViT has better adversarial robustness compared to Convolutional Neural Networks (CNNs). They find that ViT features contain low-level information that provides superior robustness against adversarial attacks. They note the combination of CNNs and transformers leads to better robustness compared to pure transformer models with increasing size or added layers.
Additionally, they find that pretraining larger datasets do not improve robustness. For a robust model, the opposite is applicable. Bhojanapalli et al.~\cite{bhojanapalli2021understanding} investigated various measures of the robustness of ViT models and resnet models against adversarial examples, natural examples, and common corruptions. The authors have investigated robustness to perturbation both to the input and to the model. It is observed that transformers are robust to remove any single layer from either the input or the model.

Paul et al.~\cite{paul2022vision} studied various aspects of robust learning methods of ViT ~\cite{dosoViTskiy2020image}, CNNs, and Big transformer ~\cite{kolesnikov2020big}  methods. Paul et al.~\cite{paul2022vision} benchmarked the robustness of ViTs on a wide range of ImageNet datasets. Their results are in table-\ref{tab:my_label}. Through six experiments, the authors verified that ViT has improved in robustness compared to CNN and BIG transformers. The results of those experiments include Experiment 1: attention is crucial for improved robustness; Experiment 2: the important role of pretraining; Experiment 3: ViT has better robustness to image masking; Experiment 4: Fourier spectrum analysis reveals low sensitivity for ViT; Experiment 5: adversarial perturbation has spread wider in the energy spectrum; and Experiment 6: ViT has smoother Loss Landscape to input perturbations. Hendrycks et al. \cite{hendrycks2018benchmarking} has been introduced for benchmarking neural network robustness against common corruptions like  natural occurring and corruptions using ImageNet-C ~\cite{hendrycks2018benchmarking} dataset. The ImageNet-C dataset contains perturb version of the original ImageNet dataset. There are 1000 classes and each has 50 images. The performance of various models is shown in the table-\ref{tab_robust_2}.

ViT ~\cite{dosoViTskiy2020image} models are less effective at capturing the high-frequency component of images as compared to CNNs, as investigated by Park et al. ~\cite{park2021vision}. HAT ~\cite{bai2022improving} was the result of a further investigation into the effect of an existing transformer model from a frequency perspective. HAT perturbs the high-frequency component of the input image with noise using the RandAugment method. Wu et al. ~\cite{wu2022towards} investigated the issue of transformer models vulnerable to adversarial examples like CNNs. This issue is handled in CNNs with the help of adversarial training, which is the most effective way to accomplish it in CNNs. But in transformers, the adversarial training has a heavy computational cost due to the quadratic complexity of the self-attention computation. The AGAT method uses an efficient attention-guided adversarial mechanism with removing certainty patch embedding on each layer with an attention-guided dropping strategy during adversarial training. Bai et al.~\cite{bai2022improving} have proposed the HAT method (named for High-frequency components via Adversarial Training), which perturbs high-frequency components during the training stage. The HAT method alters the high-frequency components of the training image by adding adversarial perturbation and then trains the Vision Transformer (ViT) ~\cite{bai2022improving} model with the altered image to improve the model performance and makes the model more robust.

\subsection{Privacy} 
Today, pre-trained transformer models are deployed on cloud systems. One of the main issues in cloud-based model deployment pertains to privacy issues in the data. The major privacy issues are the exposure of user data such as search history, medical records, and bank accounts. The current research focuses  on preserving privacy in the inference of transformer models. The paper~\cite{huang2020texthide} introduced TextHide, a federated learning technique to preserve privacy, but this method is for sentence-based like machine translation, sentiment analysis, and paraphrase generation tasks), rather than for token-based tasks (such as name entity recognition and semantic role labeling).  Similarly, the  DP-finetune~\cite{kerrigan2020differentially} Differential Privacy (DP) method allows us to quantify the degree to which we can protect the sensitivity of data. But, training a DP algorithm degrades the quality of the model, which can be tuned using a public base model on a private dataset.  

The paper~\cite{chen2022x} proposed THE-X as a method by series of approximations on the HE~\cite{boemer2020mp2ml} based solution in a transformer. THE-X method replaces non-polynomial operations with a series of approximations with the help of these layers such as the SoftMax and the GELU layer, drop the pooler layer, add Layer normalization, use knowledge distillation techniques, and then use HE-supported operations with HE transformer. THE-X method is evaluated using BERT-Tiny Model on GLUE~\cite{wang2018glue} and benchmarked for a CONLL2003~\cite{sang2003introduction} task.    


\subsection{Approximation} 
In this section, we consider various kinds of differential equations and their approximation methods to make them more efficient in terms of computational cost and number of parameters. The paper~\cite{ruthotto2020deep} was one of the first to provide a theoretical foundation based on Partial Differential Equations (PDEs) for deep neural networks such as ResNets. More specifically, the author showed that residual CNNs could be interpreted as a discretization of a space-time differential equation. Based on the theoretical characterization, Ruthotto also proposes new models such as hyperbolic and parabolic CNNs with special properties. Residual networks have also been interpreted as Euler discretizations of Ordinary Differential Equations (ODEs). However, the Euler method of solving is not precise and has truncation errors as it is a first-order method. The authors of ODE Transformers ~\cite{li2021ode} used a classical higher-order method (Runge Kutta) to build a transformer block. They evaluated the ODE transformer on three sequence-generation tasks. These tasks proved the transformer's effectiveness, including abstractive summarization, machine translation, and grammar error correction. Another effort in this direction is TransEvolve ~\cite{dutta2021redesigning}, which provides a Transformer architecture such as ODE transformer but is modeled on multi-particle dynamic systems.

Transformers have been shown to be equivalent to universal computation engines ~\cite{lu2021pretrained}. The authors have proposed an architecture known as the Frozen Pretrained Transformer (FPT), which can be trained on a single modality (such as text data for language modeling), and identify abstractions (such as feature representations) that are useful across modalities. They have taken a GPT, pre-trained it on only natural language data, and fine-tuned its input and output layers along with the layer normalization parameters and positional embeddings. This has resulted in the FPT performing comparably with transformers trained completely from scratch for a variety of tasks such as protein fold prediction, numerical computation, and even image classification.

 \begin{table*}
\centering
\begin{tabular}{p{7em}p{9em}p{3em}p{2em}p{4em}p{19em}}
\hline
Tasks & Corpus & Length &  Class &  Metrics& Description\\
\hline\hline

 Long LISTOPS & ListOps~\cite{nangia2018listops} & 2K & 10 & Accuracy & To check the ability to reason hierarchically while handling long contexts\\
 
 Byte-level Text Classification & IMDB Reviews~\cite{maas2011learning} & 4K & 2 &Accuracy  & Is the model the  classify the text  associated with the real-world applications? \\

 Byte-level Document Retrieval& AAN~\cite{radev2013acl} & 8K & 2 & Accuracy & Does the model able to retrieve the matched sequence document 1 ?\\ \hline
 Image Classification& CIFAR10~\cite{krizhevsky2009learning} &$N^2$ & 10 & Accuracy  & Does the model able classify Image of long sequence length $N^2$ pixels?\\
 
Pathfinder & Synthetic~\cite{linsley2018learning} & 1K & 2&  Accuracy& Is there a connected path exist between point-1 and point-2 of image size 32 X 32?\\ 

 Path-X & Synthetic~\cite{linsley2018learning} & 16K & 2 &  Accuracy & Is there a connected path exist between point-1 and point-2 of image size 128 X 128? \\

\hline
\end{tabular}
\caption{This table reports a description of  various tasks in the Long Range Arena (LRA) benchmark. Along with the description, it provides corpus details, length of the sequence, number of classes, and evaluation metrics for all the tasks in the LRA benchmark.  }\label{LRA_benchmark_info}
\vspace{-0.7em}
\end{table*}

\subsection{Probabilistic methods}
The bayesian-based probabilistic model plays an important role in estimating uncertainty in data and model while classifying an image. It brings efficiency in terms of less number of parameters (by sampling models using dropout) and cab able to model efficiently using a small amount of data. Guo et al.~\cite{guo2022uncertainty} have proposed Uncertainty-Guided Probabilistic Transformer (UGPT) for Complex Action Recognition. Multi-head self-attention is used to capture the complex and long-term dynamics of complex actions. The author has extended the deterministic transformer mechanism to the probabilistic transformer mechanism to quantify the prediction's uncertainty. The author introduces a novel training strategy using majority and minority models for estimating epistemic (model) uncertainty. Yang et al. ~\cite{yang2021uncertainty} have discussed the difficulty in camouflaged object detection due to indistinguishable textures leading to inherent uncertainty on it. The author proposed an uncertainty-guided transformer reasoning (UGTR) to learn a conditional distribution over the model output to estimate uncertainty and make reasoning over it.

\subsection{Efficient learning: Continual/ Incremental/ Lifelong/ and Federated learning } 
Another important aspect is how we will adapt the trained model with a few new categories of data and new tasks. In recent years much study has been done in the field of continual learning/incremental learning/ lifelong learning to take care of new classes and new tasks etc. our scope is not to define these terms as per definition. Here we focus on how the transformer model helps to improve these learning processes.
Dytox~\cite{douillard2022dytox} has discussed transformer models for continual learning with dynamic token expansion. The author has discussed the issue of the existing deep method's struggle to continually learn new tasks without forgetting the previous ones. The transformer's encoder and decoder modules are shared among all tasks in order to scale it for a large number of tokens. The author has introduced a lifelong vision transformer (LVT) ~\cite{Wang_2022_CVPR} to get rid of catastrophic forgetting in Continual Learning tasks. In order to achieve this, the author has introduced an attention-based mechanism to get better stability for continual learning. Wang et al. ~\cite{wang2022online} have introduced a contrastive vision transformer to mitigate catastrophic forgetting. The method uses a contrastive learning strategy using a transformer to get a better stability-plasticity trade-off for continual online learning.





\subsection{Inclusiveness}
The Inclusiveness domain focus on the 
The transformer model-based research focuses on empowering everyone and engaging people with visual, hearing, and other impairments. We define this as  Inclusiveness. The main challenge is how we will deploy the transformer model in embedding systems(Like Microcontroller and Microprocessor) more efficiently for general proposed applications and also an application for impaired persons. Very little research has been carried out in this field like ViT Cane~\cite{kumar2021ViT},  Flying guide dog~\cite{tan2021flying}, TransDARC~\cite{peng2022transdarc}, and Trans4Trans~\cite{zhang2021trans4trans}. ViT Cane~\cite{kumar2021ViT} provide visual assistants like finding the shortest path to a destination and detecting obstacle from a distance for visually impaired persons. ViT can model uses a PI camera module to capture the picture and detect the obstacle using the ViT transformer model. The complete work is done using a Raspberry Pi microcontroller. The author has compared the method with CNN-based TOLO architecture and shows better results. Flying guide dog~\cite{tan2021flying} helps to discover the walkable path for a visually impaired person. They use drones to capture real-time street views and a transformer model to extract the walkable area from the segmentation predictions. Finally, the drone adjusts the movement automatically and guides the person to walk in the walkable areas. Trans4Trans~\cite{zhang2021trans4trans}  uses an efficient transformer for transparent object detection and semantic scene segmentation in real-world navigation assistance.
TransDARC~\cite{peng2022transdarc} uses a transformer-based model for Driver ActiViTy Recognition using latent space feature calibration. The author observes the slowness in understanding the driver's behavior and response time of the existing model compared to the transformer-based model. We account for efficiency in terms of comparing performance transformer models with existing CNN-based models in terms of speed in ViT-Cane and other parameters in the above papers. We also call it the efficient way of deploying transformer models on embedded systems.


 \begin{table*}[htb]
\centering
\begin{tabular}{p{7em}p{3em}p{3em}p{3em}p{3em}p{3em}p{3em}p{3em}} 
\hline
Model & ListOps & Text & Retrieval  & Image & Pathfinder & Path-X & Avg \\
\hline\hline
Chance &10.00& 50.00& 50.00& 10.00& 50.00& 50.00& 44.00\\
Transformer& 36.37& 64.27& 57.46& 42.44& 71.40& FAIL& 54.39\\ \hline
Local Attention& 15.82& 52.98& 53.39& 41.46& 66.63& FAIL& 46.06\\
Sparse Trans.& 17.07& 63.58& \textbf{59.59}& 44.24& 71.71& FAIL& 51.24\\
Longformer& 35.63& 62.85& 56.89& 42.22& 69.71& FAIL& 53.46\\
Linformer& 35.70& 53.94& 52.27& 38.56& 76.34& FAIL& 51.36\\
Reformer& \textbf{37.27}& 56.10& 53.40& 38.07& 68.50& FAIL& 50.67\\
Sinkhorn Trans.& 33.67& 61.20& 53.83& 41.23& 67.45& FAIL& 51.39\\
Synthesizer& 36.99& 61.68& 54.67& 41.61& 69.45& FAIL& 52.88\\
BigBird& 36.05& 64.02& 59.29& 40.83& 74.87& FAIL& \textbf{55.01}\\
Linear Trans.& 16.13& \textbf{65.90}& 53.09& 42.34& 75.30& FAIL& 50.55\\
Performer& 18.01& 65.40& 53.82& \textbf{42.77}&\textbf{77.05}& FAIL& 51.41\\ \hline
Nyströmformer* & 37.34& 65.75& 81.29& -&-& -& 61.46\\
Transformer-LS* & 38.36& 68.40& 81.95& -&-& -& 62.90\\
\hline

\hline
\end{tabular}
\caption{ Long Range Aerna results for 6 different tasks for various efficient transformer models. * tends for evaluation as per the proposed transformer  paper}\label{EVT_tab_3}
 \vspace{-0.9em}
\end{table*}

 
 \section{Dataset, and Evaluation }\label{sec:eval}
 In this section, we have discussed various data sets for the Image classification task and evaluated the models' performance on those datasets. We have included three important profiling areas for efficient transformers such as 1. State of the Art comparison on the ImageNet dataset, 2. Transfer learning on new datasets, and 3. Long Rage Arena (LRA) performance.
\subsection{Dataset}
We compare various transformer results for image classification tasks on ImageNet-1K~\cite{deng2009imagenet} dataset and ImageNet-21K~\cite{deng2009imagenet}. The ImageNet-1K dataset contains 1.2 million images in the training set and 50K images in the test set over the 1000 (1K) category of classes. ImageNet-21K is a large-scale dataset containing 14 million images in the training set over the 21K category of classes. We show the transfer learning performance from ImageNet-1K to other datasets like CIFAR10~\cite{krizhevsky2009learning}, CIFAR100~\cite{krizhevsky2009learning}, Oxford-IIIT-Pet~\cite{parkhi2012cats}, Oxford-IIIT-Flower~\cite{nilsback2008automated} and Standford Cars~\cite{krause20133d}. We show transfer learning performance from ImageNet-21K to the above datasets in table-\ref{EVT_tab_2}. We report the results transformer model on ImageNet-C~\cite{hendrycks2018benchmarking} dataset. We have discussed another benchmark dataset for transformers models such as Long Rage Arena~\cite{tay2020long} (LRA), which consists of six challenging corpora focused on long-range sequences.

\subsection{State of the art comparison  on ImageNet Dataset}\label{model_performance}
We analyzed and compared the model performance of various efficient vision transformers on ImageNet-1k~\cite{deng2009imagenet} dataset as shown in the table-~\ref{EVT_tab_1}. The comparison is based on the number of parameters in millions (M), number of floating point operations (FLOPS), image size, type of networks, and top-1 accuracy. In table-~\ref{EVT_tab_1}, table-~\ref{EVT_tab_2_extention}, and table-~\ref{EVT_tab_1_extention} we compare various transformer's performance for input size- $224 \times 224$ pixels, whereas in table-\ref{EVT_tab_1_1}, we compare with different image sizes like $256 ^2$, $288 ^2$, $384 ^2$, $448 ^2$, $512 ^2$, and $600 ^2$. In table-\ref{EVT_tab_1}, we start comparing with Convolution Neural Net (CNN) architectures like ResNets~\cite{he2016deep} and RegNet~\cite{radosavovic2020designing} provide good performance on ImageNet ~\cite{deng2009imagenet} dataset, ImageNet Real Dataset  ~\cite{beyer2020we} and ImageNet-v2~\cite{recht2019imagenet} dataset. Similarly, Involution Neural Network (INN)~\cite{li2021involution} provides good performance on ImageNet data compared to ResNet models. We start comparing transformer base architecture DeiT~\cite{touvron2021training} with both CNN and INN-based architecture for image classification tasks and observe that transformer architecture performs better compared to ResNet and RegNet architecture. 

In comparison with all efficient transformers, WaveViT~\cite{yao2022wave} transformer model is more efficient as compared to the number of parameters  and its top-1 accuracy. WaveViT-$S^*$  performs better with small parameters(22.7M) and has a comparative performance(83.9 top-1 accuracy) with other models on the ImageNet-1k dataset and WaveViT-$L^*$ provides the best performance  in terms of top-1 accuracy (85.5\% with 57.5M of parameters) among all the transformer-based models. But the WaveViT model uses supervised extra training data to achieve this performance. CvT~\cite{wu2021CvT} model has been evaluated on ImageNet, ImageNet real ~\cite{beyer2020we} dataset, and ImageNet-v2~\cite{recht2019imagenet} dataset and Performs comparative results on all three datasets with a comparatively small number of parameters (32M) for the image of size 224x224 pixels. We analyzed top-1 accuracy for images of size 384x384. We found that CMT ~\cite{guo2022cmt} with a small number of parameters(25.1M) and less number of FLOPS(4G) provides an equivalent top-1 score(83.3) on ImageNet-1k dataset and also evaluated on ImageNet real and ImageNetV2~\cite{recht2019imagenet} dataset as well.  But Uniformer~\cite{li2022uniformer} performs best among all the models for image size $224^2$ with more parameters (100M)and FLOPs(12.6) as shown in table-~\ref{EVT_tab_2_extention}. Similarly, we provide comparison of various MLP-like transformer models as in table-~\ref{EVT_tab_1_extention}.  we report that Hire-MLP-Large~\cite{Guo_2022_CVPR}  model provides the best performance  in terms of top-1 accuracy (83.8\% with 96M of parameters) among all the MLP-Mixer type transformer models. Similarly, in the pooler type network, we report that PoolFormer-M48~\cite{yu2022metaformer}  model provides the best performance  in terms of top-1 accuracy (82.5\% with 73M of parameters) among all pooler type transformer models.

Similarly, table-\ref{EVT_tab_1_1} reports the performance of various transformer models in various image sizes. For image size $256^2$, CMT-B\cite{guo2022cmt} performs better, whereas, for image size $288^2$, LV-ViT~\cite{jiang2021all} performs best, but it uses extra labeled data to train the model. Similarly, for image size $384^2$, CSwin-B\cite{dong2022cswin} performs best(85.4\% with 78M of parameters and 47G number of FLOPs) for transformer type networks and Uniformer\cite{li2022uniformer} performs good (86.3\% with 100M of parameters and 39.2G number of FLOPs), whereas MaxViT\cite{tu2022maxvit} performs best (86.4\% with 212M of parameters and 133.1G number of FLOPs) for convolution transformer type networks. For image size $448^2$, CaiT\cite{touvron2021going} performs best(86.3\% with 271M of parameters and 247.8G number of FLOPs) for transformer type networks. For image size $512^2$, LV-ViT\cite{jiang2021all} performs well (86.4\% with 151M of parameters and 214.8G number of FLOPs), whereas MaxViT\cite{tu2022maxvit} performs best (86.7\% with 212M of parameters and 245.4G number of FLOPs),  for convolution transformer-type networks with extra labeled trained data.

Table-\ref{EVT_tab_5} shows the performance analysis of spectral transformer models trained on ImageNet1K~\cite{deng2009imagenet} for image size $224 \times 224$ and $384 \times 384$. It compares the number of parameters, number of FLOPs, and Top-1 accuracy of various spectral vision transformers. The green tick mark in the WaveViT~\cite{yao2022wave} model indicates attention is used  in the transformer model. The green tick mark in the extra level column indicates that WaveViT~\cite{yao2022wave} uses extra training data while training. Fnet~\cite{lee2021fnet}, GFNet~\cite{rao2021global} and AFNO~\cite{guibas2021efficient} do not use self-attention network compared to WaveViT~\cite{yao2022wave} and achieve good performance  with less number of parameters and FLOPs compare to WaveViT~\cite{yao2022wave}. Only WaveViT~\cite{yao2022wave} uses extra training data during the training of the transformer models. It is observed that the extra data helps to improve the performance in the ImageNet-1K dataset.

Similarly, we report the performance of various transformers on the ImageNet-21K dataset as shown in the table-~\ref{EVT_tab_1_2}. Here we compare types of transformer networks like a transformer and Convolution transformer, Number of Parameters, Number of FLOPs, image sizes, and top-1 accuracy.

\subsection{Transfer learning  on New datasets for Image Classification task}
Table-~\ref{EVT_tab_2}, shows the transfer learning capability of the pre-trained transformer models on CIFAR10~\cite{krizhevsky2009learning}, CIFAR100~\cite{krizhevsky2009learning}, Pets~\cite{parkhi2012cats}, Flowers~\cite{nilsback2008automated}, Cars~\cite{krause20133d} datasets. We show the comparison of the number of parameters, number of FLOPs, and top-1 accuracy of various models on the above datasets. The table~\ref{EVT_tab_2} contains two blocks, the top blocks show a comparison of the transformer models trained on the ImageNet-1k~\cite{deng2009imagenet} (1000 target categories) dataset, whereas the bottom block shows the comparison of the transformer models on ImageNet-22K ~\cite{deng2009imagenet}(22000 target categories). 
In the top block, we observed that CMT-S~\cite{guo2022cmt} performs better in CIFAR10~\cite{krizhevsky2009learning} dataset i.e.,  accuracy is around 99.2\%, RegionViT-M~\cite{chen2022regionViT}, DeiT-B~\cite{touvron2021training},CaiT-S~\cite{touvron2021going} performs better in CIFAR100~\cite{krizhevsky2009learning} dataset i.e.,  accuracy is around 90.8\%. In  Oxford-IIIT-Pet ~\cite{parkhi2012cats} dataset,  RegionViT-M\cite{chen2022regionViT} performs better and its accuracy is around 95.5\%. GFNet-H-B\cite{rao2021global} and TNT-S~\cite{han2021transformer} performs better in Oxford-IIIT-Flowers~\cite{nilsback2008automated} dataset i.e.,  accuracy is around 98.8\%.
In Standford Cars~\cite{krause20133d} dataset,CMT-S~\cite{guo2022cmt} performs better and its accuracy is  94.4\%. 

In bottom block, we observe that CvT-W24~\cite{wu2021CvT} performs better in CIFAR10~\cite{krizhevsky2009learning}, CIFAR100~\cite{krizhevsky2009learning}, and Oxford-IIIT-Flowers~\cite{nilsback2008automated} datasets, whereas ViT-H~\cite{dosoViTskiy2020image} performs well on the Oxford-IIIT-Pet ~\cite{parkhi2012cats} dataset. We also analyzed model architecture across layers for ViT~\cite{dosoViTskiy2020image} and CvT~\cite{wu2021CvT} models. We observe that ViT-H with 16 patch size performs better than ViT-Base and ViT-Large.  A similar case with CvT-24 performs better compared to CvT-13 and CvT-21. So we can conclude that the larger model size performs better.

It is very difficult to claim that pre-training models on  ImageNet-22K ~\cite{deng2009imagenet} dataset gives better representation features, which helps to perform better in Transfer learning as compared to pre-training models on  ImageNet-1K~\cite{deng2009imagenet} because the models are different. It is not clear if it is due to the model, or it is due to the large dataset with more classes.

\subsection{Long Range arena (LRA) Benchmark}
Transformers are largely not performing very well on long sequence lengths due to quadratic complexity in self-attention. Long Range Arena (LRA)~\cite{tay2020long} is another evaluation benchmark method focused on evaluating model quality on long-range- context scenarios. LRA benchmark is applicable to sequence lengths ranging from 1K to 16K tokens. It focuses on a wide range of data types and modalities, such as mathematical reasoning requiring similarity, structural, text, natural, synthetic images, and  visual-spatial reasoning. LRA benchmark is basically evaluated the efficiency of transformer models on a list of tasks focus on long-range data contexts such as Long ListOps task, Byte level Text Classification, Byte Level Document Retrieval task,  Image classification on the sequence of pixels, Pathfinder and Pathfinder-X task as shown in table-~\ref{LRA_benchmark_info}. LRA benchmark is created based on the Generality of the model(the efficient transformer model should apply to a variety of the task), Simplicity(the task should have simple to setup), long inputs (the input sequence length should have reasonably long to capture long-range dependency of the model), Challenging(the task should have difficult enough for improvement in the model performance and to encourage future research direction),  Probing diverse aspects ( the set tasks should access the different capabilities of the model like hierarchical/spatial structures, generalizations capabilities, etc.) and Non-resource intensive(the model should be designed to be the lightweight model which can be accessible to researcher).

 The LRA benchmark is evaluated on recent transformer models such as Performer~\cite{choromanski2020rethinking}, Reformers~\cite{kitaev2019reformer}, Linformers~\cite{wang2020linformer}, Linear transformers~\cite{katharopoulos2020transformers}, Synthesizers~\cite{tay2021synthesizer}, Sinkhorn transformer~\cite{tay2020sparse}, Sparse transformers~\cite{child2019generating},  Nystromformer~\cite{xiong2021nystromformer}, Transformer-LS~\cite{zhu2021long}, Longformers~\cite{beltagy2020longformer} and Big bird~\cite{zaheer2020big}. None of the latest vision transformer(like Swin~\cite{liu2021swin}, Twin~\cite{chu2021twins}, CvT~\cite{wu2021CvT}, CSwin~\cite{dong2022cswin}, RegionViT~\cite{chen2022regionViT}, WaveViT~\cite{yao2022wave}) models are not evaluated on this benchmark. It is more challenging and interesting to see results on this benchmark.  Quantitative results for various transformer models are reported in table-\ref{EVT_tab_3}. From the table, we observe that it is a very challenging benchmark. The performance in visual domains is relatively low compared to language tasks. The image classification scores low for long sequences, and Most of the time, the model fails in the Path-X task.

 \section{Conclusion}
 We have discussed all ten dimensions of the efficient 360 frameworks in different subsections. The survey has opened up new research areas and directions for transformers. For instance, we are ourselves coming up with a spectral neural operator-based transformer which we believe is likely to outperform state-of-art transformers with respect to robustness, explainability, and efficiency (have a significantly lesser number of parameters).

 We also notice from the diagram (Figure 1), there is room for research in dimensions such as privacy, transparency, fairness, efficient learning, and most importantly inclusiveness. Due to the applicability of some of the advanced transformer models on other modalities of data, a lot of work on audio, speech, and video has opened up. Further, AI for sciences is another open area of research including the applicability of the advanced transformer models on high-resolution data like weather forecasting and oceanography (wave modeling).

\bibliographystyle{named}
\bibliography{ijcai23_org}

\end{document}